\documentclass[ijoc,nonblindrev]{informs3} 

\SingleSpacedXII 

\usepackage{natbib}
 \bibpunct[, ]{(}{)}{,}{a}{}{,}%
 \def\bibfont{\small}%
\TheoremsNumberedThrough     

\EquationsNumberedThrough    

\MANUSCRIPTNO{00-00} 
\begin{document}



\RUNTITLE{Boost-R}


\vspace{20pt}
\TITLE{\Large{Boost-R: Gradient Boosted Trees for Recurrence Data} \vspace{20pt}}

\ARTICLEAUTHORS{%
\AUTHOR{\normalsize{Xiao Liu}}
\AFF{\normalsize{Department of Industrial Engineering, University of Arkansas}}
\AFF{\normalsize{University of Arkansas, \EMAIL{xl027@uark.edu}}}
\AFF{}
\AUTHOR{\normalsize{Rong Pan}}
\AFF{\normalsize{School of Computing, Informatics, Decision Systems Engineering}}
\AFF{\normalsize{Arizona State University, \EMAIL{rong.pan@asu.edu}}}
} 


\ABSTRACT{%
\normalsize{
\indent Recurrence data arise from multi-disciplinary domains spanning reliability, cyber security, healthcare, online retailing, etc.
This paper investigates an additive-tree-based approach, known as Boost-R (Boosting for Recurrence Data), for recurrent event data with both static and dynamic features. Boost-R constructs an ensemble of gradient boosted additive trees to estimate the cumulative intensity function of the recurrent event process, where a new tree is added to the ensemble by minimizing the regularized $L^2$ distance between the observed and predicted cumulative intensity. 
Unlike conventional regression trees, a time-dependent function is constructed by Boost-R on each tree leaf. The sum of these functions, from multiple trees, yields the ensemble estimator of the cumulative intensity. The divide-and-conquer nature of tree-based methods is appealing when hidden sub-populations exist within a heterogeneous population. The non-parametric nature of regression trees helps to avoid parametric assumptions on the complex interactions between event processes and features. Critical insights and advantages of Boost-R are investigated through comprehensive numerical examples. Datasets and computer code of Boost-R are made available on GitHub. To our best knowledge, Boost-R is the first gradient boosted additive-tree-based approach for modeling large-scale recurrent event data with both static and dynamic feature information.}}%

\maketitle

%


\clearpage
\SingleSpacedXII

\section{Introduction}
\vspace{8pt}

\subsection{Background} \label{sec:objectives}
The rapid penetration of IoT technologies gives rise to large-scale recurrent event data from multidisciplinary domains, including reliability, asset management, clinical trials, cyber security, etc. A typical recurrent event dataset has two defining characteristics: (\textbf{\textit{i}}) the event occurrence times experienced by individuals are recorded; (\textit{\textbf{ii}}) each individual is characterized by both static and dynamic feature/covariate information. By learning the relationship between event processes and features, statistical approaches are needed to better understand why critical events happened in the past, when events of interest will recur in the future, and how one could optimize the event processes through proactive interventions. 

\begin{figure}[h!]
	\begin{center}
		\includegraphics[width=1\textwidth]{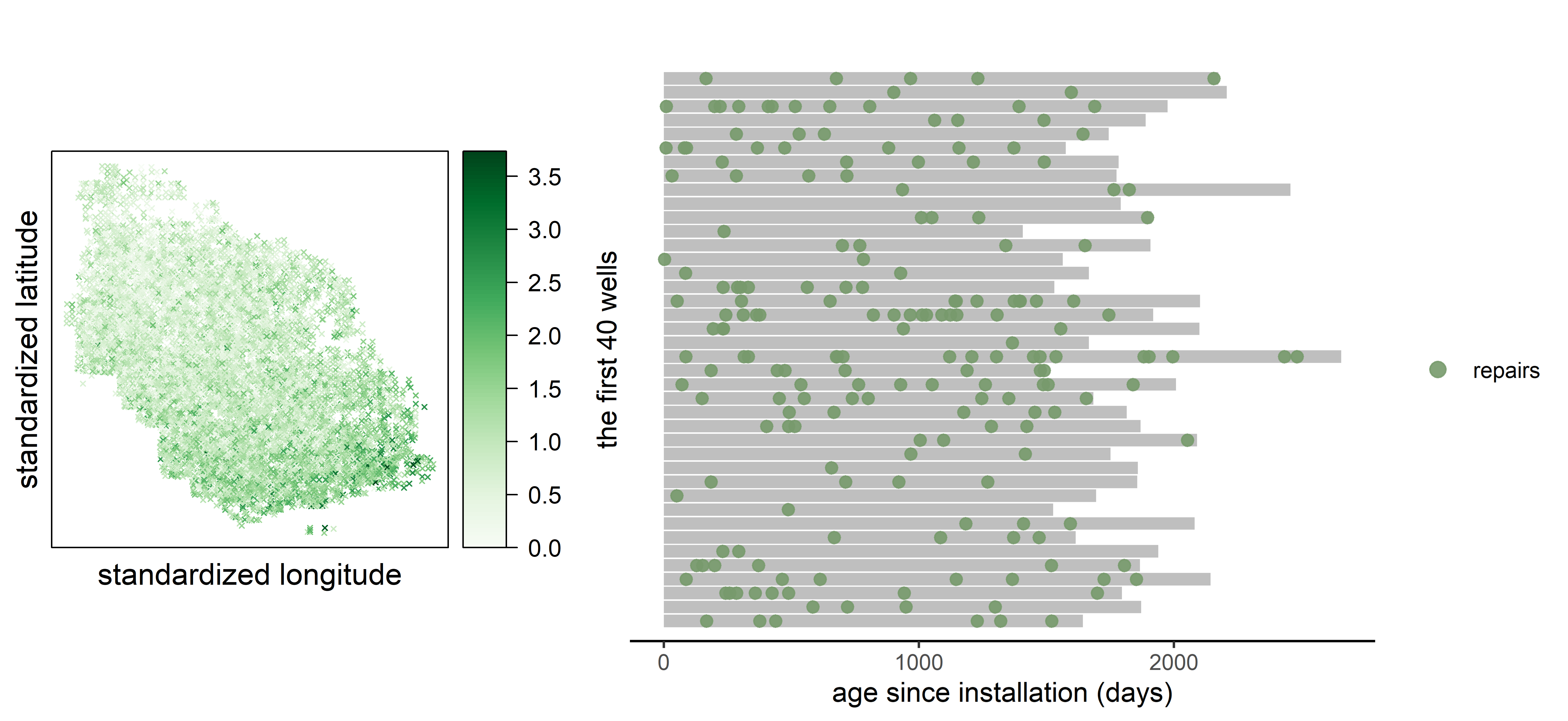}
	\end{center}
	\caption{Left panel: geo-locations of 8232 wells with the legend showing the average number of failures per year; Right panel: illustration of the failure processes of the first 40 wells.}
	\label{fig:wells} 
\end{figure}

For example, \cite{Liu2019} considered an asset management problem of a large fleet of oil and gas wells. An individual well, throughout its production life, requires repeated maintenance due to failures over time. Because oil and gas wells are often located in vast and remote spatial areas, the capabilities of predicting failures greatly facilitate maintenance planning and reduce total production cost. 
For individual wells, times of failures (i.e., repairs) are well documented. As a motivating example, the left panel of Figure \ref{fig:wells} shows the geo-locations of 8232 oil and gas wells installed between 2007 and 2017, while the right panel of Figure \ref{fig:wells} shows the failure processes of a pump-related failure mode for 40 selected wells.

%
%
%
%
%
%
%
%
%
%
%

\begin{figure}[h!]
	\begin{center}
		\includegraphics[width=1\textwidth]{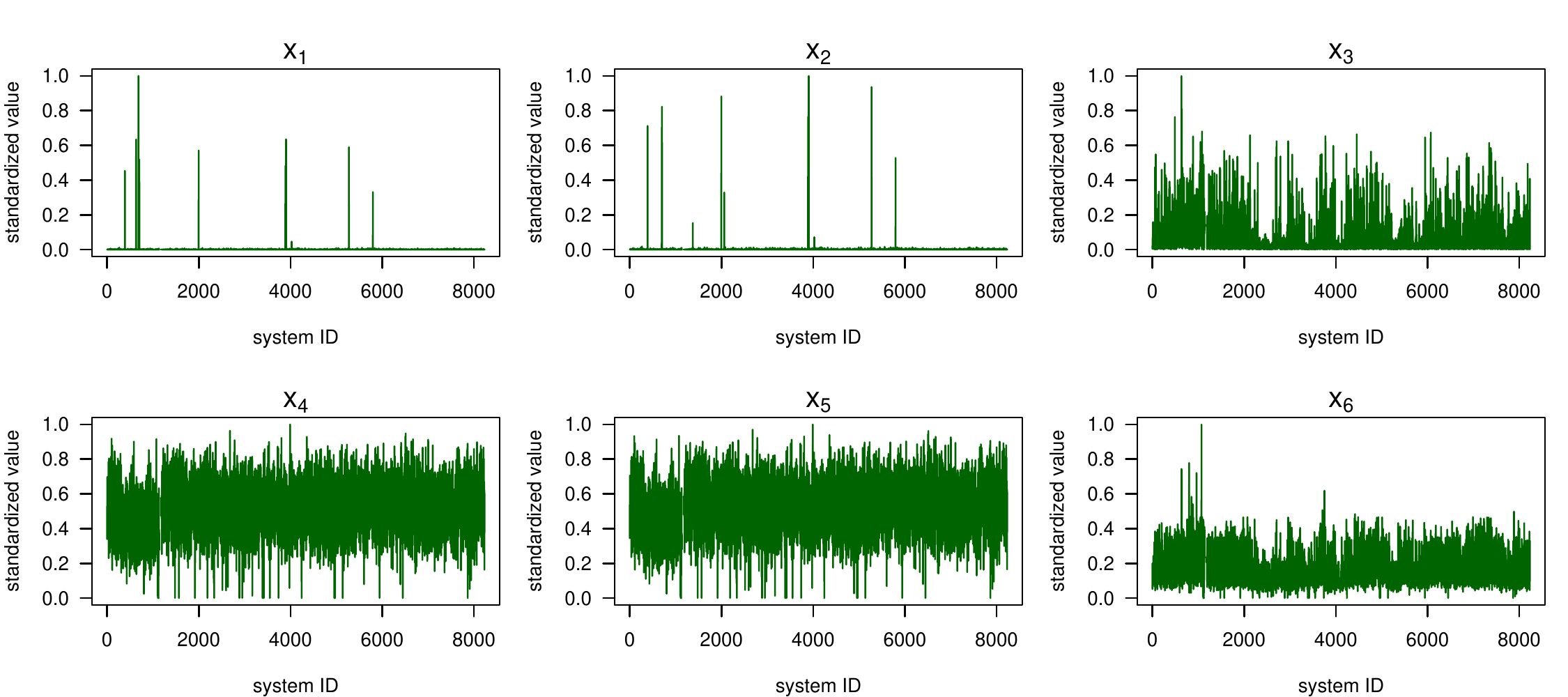}
	\end{center}
	\caption{Standardized values of $x_1,x_2,...,x_6$ for all 8232 systems.}
	\label{fig:X_case} 
\end{figure}

In this example, eight static well attributes are available, $x_1,x_2,...,x_8$. In particular, $x_1,x_2,...,x_6$ are the static well attributes, including the dimensions of two critical components ($x_1$ and $x_2$), average stroke length ($x_3$), mean-time-between-failure ($x_4$), mean polished rod horse power ($x_5$), and mean load on pump plunger against position ($x_6$). The last two attributes, $x_7$ and $x_8$, are the well latitude and longitude. Figure \ref{fig:X_case} shows the (standardized) values of $x_1,x_2,...,x_6$ for all 8232 systems. It is seen that these systems have different static well attributes. 

\begin{figure}[h!]
	\begin{center}
		\includegraphics[width=1\textwidth]{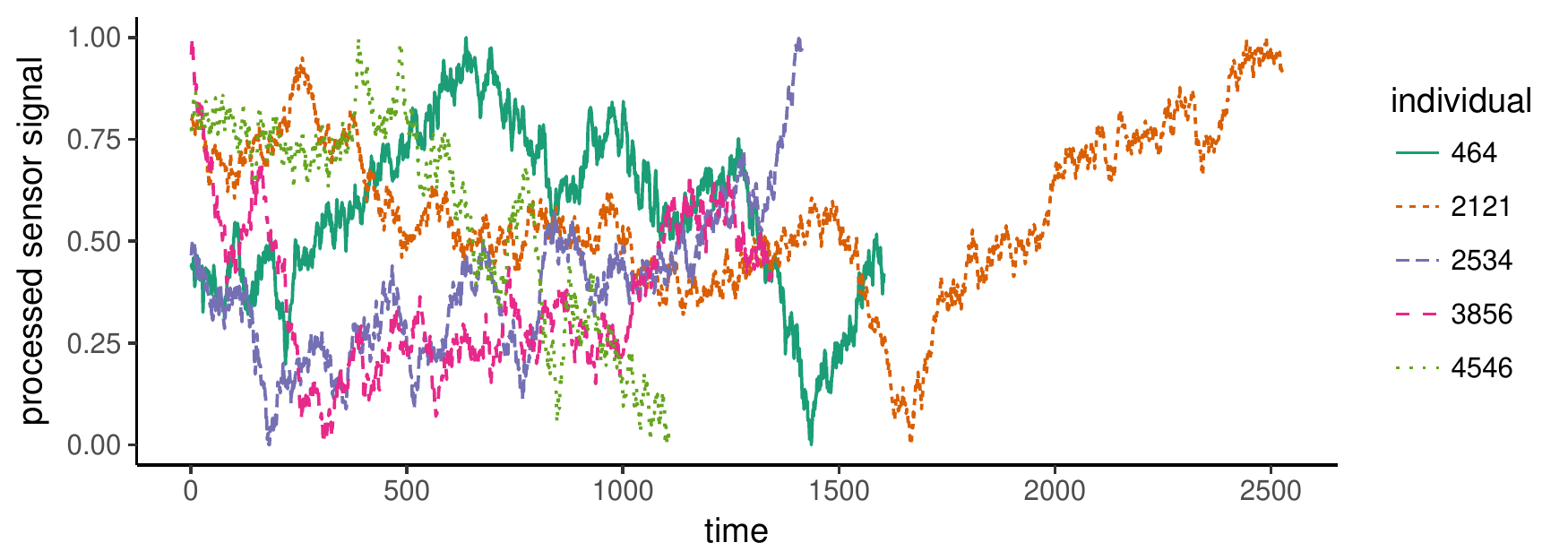}
	\end{center}
	\caption{Gear torque for 5 well systems}
	\label{fig:z} 
\end{figure}

In addition to the static well attributes, critical well operating conditions are monitored by sensors, including gear box torque, stroke length, polished rod horse power, peak surface load, pump card area and cycles. 
For illustrative purposes, Figure \ref{fig:z} shows the (standardized) gear torque for five chosen systems. 
It is seen from Figure \ref{fig:z} that these systems experience different operating conditions, which are rarely synchronized and lead to further heterogeneity in how wells fail over time.  

More application examples of recurrent event data can be found from other application domains. In personalised healthcare, it is crucial to understand the interaction between the recurrence of a chronic medical condition and a patient's risk factor; In cyber security, of interest are often the prediction of recurrent fraudulent activities on susceptible systems within a dynamic cyber environment; In transportation safety, it is important to reveal the connection between the recurrences of accidents and the traffic/visibility/weather patterns on particular highway sections; In online retail business, of interest are often the times when customers, with diverse background, re-visit the online platforms and place their orders; In video streaming quality control, of interest are often the modeling of the recurrent buffering processes during video streaming in a highly dynamic internet environment, and so on.

\vspace{8pt}
\subsection{Literature, Gaps and Contributions} \label{sec:contribution}


Statistical methods for event data have been investigated in survival analysis, reliability engineering and bioinformatics \citep{Fleming1991, Anderson1993, Nelson1995, Meeker1998a}. In \cite{Liu2019}, the authors provided a comprehensive discussions on the common challenges for modeling recurrence data from a large population of heterogeneous individuals with diverse feature information. These challenges include model specification, data heterogeneity due to different operating conditions, between-individual variation (e.g., the relationship between event process and feature information may vary among individuals) and feature selection. Hence, the divide-and-conquer nature of tree-based methods is appealing when hidden sub-populations exist within a heterogeneous population. The non-parametric nature of regression trees also helps to avoid parametric assumptions on the complex interactions between event processes and features; see \cite{Liu2019} for detailed discussions. In fact, for time-to-event data  (i.e., non-recurrent events), the advantages of tree-based methods have been investigated in \cite{Huo2006}, \cite{Hothorn2006}, \cite{Fan2006, Fan2009}, \cite{Chipman2010}, \cite{Ishwaran2008, Ishwaran2010, Ishwaran2010b} and \cite{BouHamad2011}.


In recent years, machine learning and Deep Recurrent Neural Network (DRNN) have received considerable attention for their advantages in capturing complex non-linear event-feature interactions \citep{Ranganath2016, Lao2017, Wang2017, Katzman2018, Grob2018, Lee2018}. These methods integrate statistical survival analysis models into the framework of deep learning without specifically addressing the existence of sub-populations with the event-feature relationship varying across sub-populations. Finally, as the volume of data grows, so does the amount of noise (irrelevant features, sampling error, measurement error, etc.); see \cite{Jordan2019}. Many features could be redundant from either the statistical modeling or domain knowledge perspective \citep{Guyon2003, Reunanen2003, Yuan2005, Nilsson2007, Witten2010, Paynabar2015}.

To address the challenges above, this paper proposes an additive-tree-based method for modeling recurrent event data with static and dynamic feature information. By assuming that the event process of an individual constitutes a counting process, we seek an ensemble of binary regression trees to estimate the cumulative intensity function that fully characterizes the recurrent event process. The ensemble trees are obtained under the framework of XGBoost \citep{Chen2016}, and the proposed method is called \textbf{Boost-R} (\textbf{Boost}ing for \textbf{R}ecurrence Data). To our best knowledge, Boost-R is the first gradient boosted additive-tree-based statistical learning approach for modeling recurrence data with feature information. The R code is made available on GitHub (\url{https://github.com/dnncode/Boost-R}).

\begin{figure}[h!]  
	\begin{center}
		\includegraphics[width=1\textwidth]{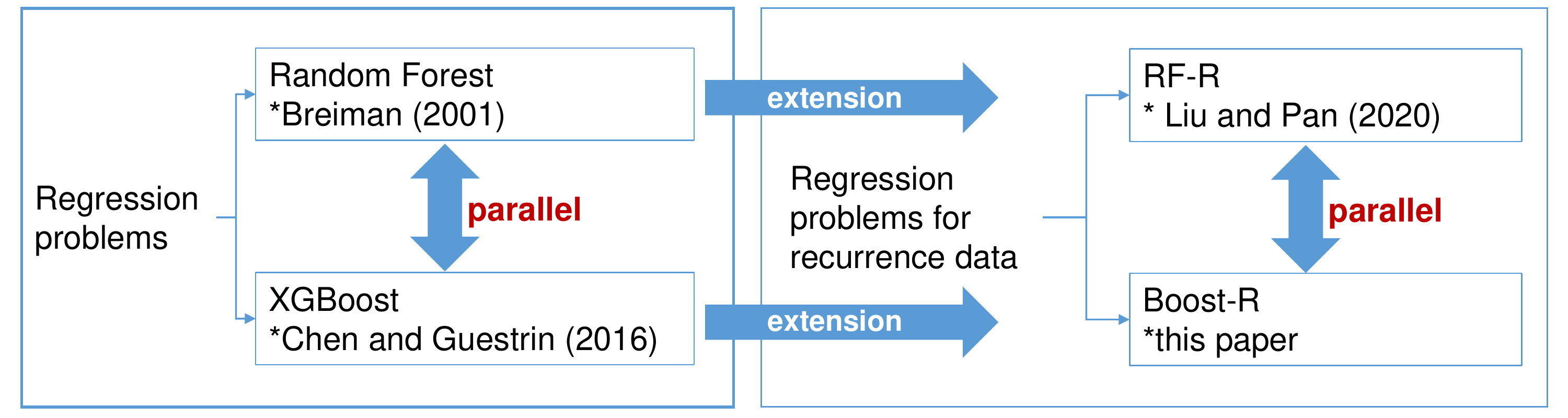}
		\centering
		\vspace{-18pt}
		\caption{The relationship between Boost-R and RF-R is similar to that between XGBoost and RF.}
		\label{fig:1}
	\end{center}
\end{figure}

Note that, a recently proposed additive-tree-based method, known as RF-R, leverages the idea of Random Forest (RF) for modeling recurrence data \citep{Liu2019}. The Boost-R proposed in this paper can be seen as a competitor of RF-R, just like how XGBoost is often seen as an alternative to RF \citep{Chipman2010, Chen2016}. The two competitors are based on exactly opposite and independent ideas: \textit{RF involves an ensemble of de-correlated and fully-grown trees, while Boosting fits a sequence of correlated simple trees (``weaker learners'') and each tree explains only a small amount of variation not captured by previous trees} \citep{Freund1997, Chipman2010, Hastie2009}.
From this perspective, a natural and meaningful question to be answered is whether the gradient boosted trees can be leveraged to achieve a better modeling and prediction performance for recurrent event data. As shown in Figure \ref{fig:1}, the relationship between Boost-R and RF-R is parallel, and the development of Boost-R in the present paper makes both RF-R and Boost-R available in practice; just like how RF and XGBoost co-exist as the two most popular off-the-shelf methods. 

\vspace{8pt}
The paper is organized as follows. 
Section \ref{sec:Boost-R} presents the technical details of Boost-R. Comprehensive numerical examples, including a case study, are provided in Section \ref{sec:numerical} which generate critical insights on the Boost-R algorithm and demonstrate its advantages. Section \ref{sec:conclusions} concludes the paper. The paper also includes supplementary material that provides additional discussions and comparison studies using different application examples. 


\vspace{16pt}
\section{Boost-R: Boosting for Recurrence Data}
\label{sec:Boost-R}

\vspace{8pt}
\subsection{The Problem Setup} \label{sec:problem}

Consider a population of $n$ individuals. An individual $i$ ($i=1,2,...,n$) experiences a sequence of $r_i$ events at times $\bm{y}_i=(y_{i,1},...,y_{i,r_i},c_{i})$, where $y_{i,1},...,y_{i,r_i}$ are the event times and $c_{i}$ is the right censoring time. 
Associated with individual $i$ there exists $p$ static features, $\bm{x}_i=(x_{i,1},x_{i,2},...,x_{i,p})$, and  $q$ time-varying dynamic features represented by a $q$-dimensional time series, $\bm{z}_i(t)=(z_{i,1}(t), z_{i,2}(t), ...,z_{i,q}(t))$. 
Hence, the recurrent event data are denoted by 
$\mathfrak{D}=(\bm{y}_i,\bm{x}_i, \bm{z}_i(t))$, where $|\mathfrak{D}|=n$, $\bm{x}_i\in \mathbb{R}^p$, $\bm{y}_i\in \mathbb{R}^{r_i+1}$ and $\bm{z}_i(t):[0,\infty)\rightarrow \mathbb{R}^q$.
Conditioning on $\bm{x}_i$ and $\bm{z}_i(t)$, events arising from an individual constitute a counting process, $\Lambda_i(t):[0,\infty)\rightarrow \mathbb{N}^+$, with the cumulative intensity $\mu(t;\bm{x}_i,\bm{z}_i^h(t))$ where $\bm{z}_i^h(t)=\{\bm{z}_i(\tau); 0 \leq \tau \leq t\}$ is the history of the dynamic feature information associated with individual $i$ up to $t$.

We seek an additive-tree-based model with $K$ binary regression trees, $T^{(1)},T^{(2)},...,T^{(K)}$, such that
\begin{equation} \label{eq:obj} 
\hat{\mu}^{(K)}(t;\bm{x}_i,\bm{z}^h_i(t)) = \sum_{k=1}^{K}T^{(k)}(\bm{x}_i,\bm{z}^h_i(t))
\end{equation}
where $\hat{\mu}^{(K)}(t;\bm{x}_i,\bm{z}^h_i(t))$ is the ensemble estimator of the time-dependent cumulative intensity function (for individual $i$) constructed from $K$ trees. Here, the output from a tree is a time-dependent function given $\bm{x}_i$ and $\bm{z}^h_i(t)$. 

Note that, there is a \textit{major difference} between conventional binary regression trees and regression trees for recurrence data. For conventional trees, a constant is found on each tree leaf and used as the predicted value for the sub-population represented by that tree leaf. When dealing with recurrent event data, on the other hand, an individual tree performs a binary partition of the feature space, and a time-dependent function (instead of a constant) needs to be found on each tree leaf. The sum of these time-dependent functions, from multiple trees, yields the ensemble estimator of the cumulative intensity which fully characterizes the recurrent event process given feature information. This key difference requires us to devise new computationally efficient algorithms for growing the ensemble trees, $T^{(1)},T^{(2)},...,T^{(K)}$, and the idea of Gradient Boosting is leveraged in this paper. 

\vspace{8pt}
\subsection{Boost-R with Static Features} \label{sec:static}

We first consider a recurrent event data set with only static features, $\mathfrak{D}=(\bm{y}_i,\bm{x}_i)$, where $|\mathfrak{D}|=n$, $\bm{x}_i\in \mathbb{R}^p$ and $\bm{y}_i\in \mathbb{R}^{r_i+1}$. A conventional binary regression tree divides the feature space into $D$ disjoint ``rectangular'' subspaces, $R_d$, $d=1,2,...,D$, and each subspace is represented by a tree leaf. For any tree leaf $d$, a constant $\mu_d$ is found as the predicted value for individuals associated with that tree leaf. Hence, a conventional regression tree can be expressed by the linear combination of indicator functions, $T(\bm{x};\bm{\theta}) = \sum_{d=1}^{D}\mu_d I(\bm{x} \in R_d)$, where $I$ is an indicator function and $\bm{\theta}=(R_d,\mu_d)_{d=1}^{D}$. 

For recurrent event data, on the other hand, each individual $i$ contains a sequence of event times $\bm{y}_i$ rather than a single response value as in the classical setting of regression trees. Hence, a time-dependent function is established on each tree leaf. The sum of these functions (from multiple trees) yields the ensemble estimator of the cumulative intensity of the event process, conditioning on the features. 
Following this idea, we search for an additive model with $K$ trees:
\begin{equation} \label{eq:additive_tree} 
\hat{\mu}^{(K)}(t;\bm{x}) = \sum_{k=1}^{K}T^{(k)}(\bm{x}) 
 = \sum_{k=1}^{K}\sum_{d=1}^{D^{(k)}}f_d^{(k)}(t) I(\bm{x} \in R_d^{(k)}).
\end{equation}

In (\ref{eq:additive_tree}), $\hat{\mu}^{(K)}(t;\bm{x})$ is the ensemble estimator of the cumulative intensity function obtained from $K$ trees. An individual tree is given by, $T^{(k)}(\bm{x}) = \sum_{d=1}^{D^{(k)}}f_d^{(k)}(t) I(\bm{x} \in R_d^{(k)})$, where $D^{(k)}$ is the number of leaves of tree $k$ and $f_d^{(k)}(t)$ is a function of time. The space of trees is given by
$\{ T(\bm{x};\bm{\theta}) =f_{r(\bm{x})}(t)\}$ with $r$ denoting the mapping from $\bm{x}$ to a tree leaf. 

The idea of boosting involves sequentially fitting a sequence of correlated simple trees (``weaker learners''), and each tree explains a small amount of variation not explained by previous trees. Let $\tilde{\mu}_i(t)$ be the empirical Mean Cumulative Function (MCF) estimator of the cumulative intensity function for individual $i$, 
then, a set of $K$ trees, $T^{(k)}$, $k=1,...,K$, can be obtained by minimizing a regularized objective function $\mathfrak{L}^{(K)}$:
\begin{equation} \label{eq:objective}
\mathrm{minimize} \quad \mathfrak{L}^{(K)} = \sum_{i=1}^{n}l(\tilde{\mu}_i(t), \hat{\mu}^{(K)}(t;\bm{x}_i)) + \sum_{k=1}^{K}\Omega(T^{(k)}).
\end{equation}
where $l$ is a differentiable convex loss function which measures the distance between $\tilde{\mu}_i(t)$ and $\hat{\mu}^{(K)}(t;\bm{x}_i)$. 
Although other choices are possible, we consider the following loss function
\begin{equation} \label{eq:loss}
l(\tilde{\mu}_i(t), \hat{\mu}^{(K)}(t;\bm{x}_i)) = \frac{1}{2}||\tilde{\mu}_i(t)-\hat{\mu}^{(K)}(t;\bm{x}_i)||_2^2
\end{equation}
which is proportional to the squared $L^2$ distance of two time-dependent functions. 

The regularization $\Omega(\cdot)$ in (\ref{eq:objective}) controls the complexity of individual trees because the central idea of Boosting involves sequentially fitting a sequence of ``weak learners''. Hence, we consider the following regularization:
\begin{equation} \label{eq:penalty}
\Omega(T^{(k)}) = \gamma_1 D^{(k)} + \frac{1}{2} \gamma_2 \sum_{d=1}^{D^{(k)}}||f_{d}^{(k)}(t)||_2^2.
\end{equation}

The first term in (\ref{eq:penalty}) controls the depth (i.e., the number of leaves) of individual trees. From the perspective of Analysis of Variance (ANOVA), the value for $D$ reflects the level of dominant interaction effects (of the features) on the recurrent event processes \citep{Hastie2009}. Hence, regularizing the depth of a tree implies that only the important main effects and lower-order interaction effects are captured. 
The second term in (\ref{eq:penalty}) adopts the shrinkage strategy in statistical learning, and penalizes the contributions from individual trees (i.e., $f_{d}^{(k)}(t)$) to the ensemble estimate (\ref{eq:additive_tree}). 


The optimization problem (\ref{eq:objective}) is a formidable combinatorial problem, but can be solved using the stagewise gradient boosting. We first discretize $\tilde{\mu}_i(t)$, $\hat{\mu}_i^{(K)}(t)$ and $f_{d}^{(k)}(t)$ at equally spaced times $\{t_j\}_{j=1}^m$, and let
\begin{itemize}
	\item $\{\tilde{\mu}_{i,j}\}_{j=1}^m$ be the values of $\tilde{\mu}_i(t)$ at $\{t_j\}_{j=1}^m$;
	\item $\{\hat{\mu}_{i,j}^{(k-1)}\}_{j=1}^m$ be the ensemble estimates from the first $(k-1)$ trees at $\{t_j\}_{j=1}^m$;
	\item $\{f_{d,j}^{(k)}\}_{j=1}^m$ be the values of $f_{d}^{(k)}(t)$ at $\{t_j\}_{j=1}^m$. 
\end{itemize} 

Then, given the first $(k-1)$ trees in the ensemble, the $k$th tree is found by minimizing the following objective function (i.e., stagewise gradient boosting):
\begin{equation} \label{eq:GB}
\mathrm{minimize} \quad \sum_{i=1}^{n} \sum_{j=1}^{m} l(\tilde{\mu}_{i,j}, \hat{\mu}_{i,j}^{(k-1)}+f_{r(\bm{x}_i),j}^{(k)}\Delta)  +\gamma_1 D^{(k)} + \frac{1}{2}\gamma_2\sum_{d=1}^{D^{(k)}}||f_{d,\cdot}^{(k)}||_2^2 
\end{equation}
where $f_{d,\cdot}^{(k)}$ is a vector $\{f_{d,j}^{(k)}\}_{j=1}^m$, and $\Delta$ is the spacing of equally-spaced times $\{t_j\}_{j=1}^m$.

Approximating the function $l$ in (\ref{eq:GB}) by a smoother function is essential in obtaining computationally efficient gradient boosting algorithms \citep{Hastie2009}. Hence, the second-order approximation of $l$ at $\hat{\mu}_{i,j}^{(k-1)}$ leads to 
\begin{equation} \label{eq:GB2}
\begin{split}
\mathrm{minimize} \quad & \sum_{i=1}^{n}\left\{ \sum_{j=1}^{m} {l}(\tilde{\mu}_{i,j}, \hat{\mu}_{i,j}^{(k-1)})+g_{i,j}f_{r(\bm{x}_i),j}^{(k)}+\frac{1}{2}h_{i,j}(f_{r(\bm{x}_i),j}^{(k)})^2\right\}\Delta \\
&\quad\quad +\gamma_1 D^{(k)} + \frac{1}{2}\gamma_2\sum_{d=1}^{D^{(k)}}||f_{d,\cdot}^{(k)}||_2^2
\end{split}
\end{equation}
where $g_{i,j} = \partial_{\hat{\mu}_{i,j}^{(k-1)}}{l}(\tilde{\mu}_{i,j}, \hat{\mu}_{i,j}^{(k-1)})$ and  $h_{i,j} = \partial^2_{\hat{\mu}_{i,j}^{(k-1)}}{l}(\tilde{\mu}_{i,j}, \hat{\mu}_{i,j}^{(k-1)})$.

Let $\Delta \rightarrow 0$ and drop the constant term ${l}(\tilde{\mu}_{i,j}, \hat{\mu}_{i,j}^{(k-1)})$ in (\ref{eq:GB2}), we have
\begin{equation} \label{eq:GB3}
\mathrm{minimize} \quad \sum_{i=1}^{n} \int\left( g_{i}(t)f_{r(\bm{x}_i)}^{(k)}(t)+\frac{1}{2}h_{i}(t)(f_{r(\bm{x}_i)}^{(k)}(t))^2\right)dt +\gamma_1 D^{(k)} + \frac{1}{2}\gamma_2\sum_{d=1}^{D^{(k)}}||f_{d,\cdot}^{(k)}(t)||_2^2
\end{equation}
where $g_{i}(t) = \partial_{\hat{\mu}_{i}^{(k-1)}(t)}{l}(\tilde{\mu}_{i}(t), \hat{\mu}_{i}^{(k-1)}(t))$ and $h_{i}(t) = \partial^2_{\hat{\mu}_{i}^{(k-1)}(t)}{l}(\tilde{\mu}_{i}(t), \hat{\mu}_{i}^{(k-1)}(t))$. 
Hence, once the first $(k-1)$ trees have been determined, the $k$th tree can be found by solving (\ref{eq:GB3}). 

Growing the $k$th tree requires iteratively finding the optimal split feature and splitting points for each tree node. Naturally, the node splitting process can be terminated when the objective function (\ref{eq:GB3}) cannot be further reduced by splitting any of the tree nodes. 
For any given tree topology, let $I_d=\{i|r(\bm{x_i})=d\}$ be a set that contains all individuals within leaf $d$. Then, the contribution to the objective (\ref{eq:GB3}) from tree node $d$ is:
\begin{equation} \label{eq:GB4}
\int\left( g_{\cdot}(t)f_{d}(t)+\frac{1}{2}(h_{\cdot}(t)+\gamma_2)f_{d}^2(t) \right)dt + \gamma_1
\end{equation}
where $g_{\cdot}(t) = \sum_{i\in I_d}g_{i}(t)$ and $h_{\cdot}(t) = \sum_{i\in I_d}h_{i}(t)$. Here, the superscript $\cdot^{(k)}$ is dropped without causing confusion.

Given any candidate split feature and splitting point, tree node $d$ can be split into two daughter nodes. Let $I_d^{(L)}$ and $I_d^{(R)}$ respectively contain the individuals in the left and right daughter nodes of $d$, the amount of reduction of (\ref{eq:GB3}) achieved by this splitting is given by:
\begin{equation} \label{eq:gain}
\begin{split}
G_1 = & \int\left(  g_{\cdot}(t)f_{d}(t)+\frac{1}{2}(h_{\cdot}(t)+\gamma_2)f_{d}^2(t) \right)dt \\
&-\int\left(g_{\cdot}^{(L)}(t)f_{d}^{(L)}(t)+\frac{1}{2}(h_{\cdot}^{(L)}(t)+\gamma_2)(f_{d}^{(L)}(t))^2 \right)dt \\
&-\int\left(g_{\cdot}^{(R)}(t)f_{d}^{(R)}(t)+\frac{1}{2}(h_{\cdot}^{(L)}(t)+\gamma_2)(f_{d}^{(R)}(t))^2 \right)dt -\gamma_1
\end{split}
\end{equation}
where $g_{\cdot}^{(L)}(t) = \sum_{i\in I_d^{(L)}}g_{i}(t)$, $h_{\cdot}^{(L)}(t) = \sum_{i\in I_d^{(L)}}h_{i}(t)$, $g_{\cdot}^{(R)}(t) = \sum_{i\in I_d^{(R)}}g_{i}(t)$ and $h_{\cdot}^{(R)}(t) = \sum_{i\in I_d^{(R)}}h_{i}(t)$.
Hence, if $\mathrm{max}G_1 > 0$, the optimal split feature and splitting point are the ones that maximize $G_1$. If $\mathrm{max}G_1\leq 0$, no  gain can be achieved by further splitting the tree node $d$, making this node a terminal node. The tree growing process is terminated when no further node splitting is possible.


For a high-dimensional feature space with a large $p$, there is always a need arising from practice to perform feature selection. Excluding irrelevant features greatly helps to develop more accurate predictive models with improved model interpretability. By design, tree-based methods have a natural advantage in terms of feature selection. Recall that, at each tree node splitting, the optimal split variable is chosen to achieve the maximum gain in (\ref{eq:gain}). Hence, the importance of a feature can be measured as the total gain achieved by splitting tree nodes based on this feature. 

To make this idea formal, for any tree $k$ in the ensemble, let $G_{d',i}^{(k)}$ be the gain achieved by splitting an internal node $d'$ by feature $i$. Note that, $d'=1,2,\cdots,D^{(k)}-1$ where $D^{(k)}-1$ is the number of internal nodes of a binary tree with $D^{(k)}$ leaves. Let $s_{d',i}^{(k)}=1$ if the internal node $d'$ is split by feature $i$; otherwise $s_{d',i}^{(k)}=0$. Then, the importance of feature $i$ can be computed as \citep{Hastie2009}:
\begin{equation}
w_{i} = \frac{1}{K^2}\sum_{k=1}^{K} \sum_{d'}^{D^{(k)}-1}G_{d',i}^{(k)}s_{d',i}^{(k)}, 
\quad\quad \text{for }i=1,2,\cdots,p.
\label{eq:importance}
\end{equation} 
If we let $G_{d',i}^{(k)}=1$, the importance of feature $i$ is measured by the number of times this feature is used for splitting a node, and such a strategy appeared in \cite{Chipman2010}. 

Finally, the Boost-R algorithm is summarized in Algorithm 1. 

\vspace{8pt}

\SingleSpacedXII
\begin{algorithm}[H]
	
	\KwData{ ($\bm{x}_i,\bm{y}_i$) for $i=1,2,...,n$}
	
	\vspace{0.2cm}
	For all $i=1,...,n$:

	initialize $\mu^{(0)}(t)=0$,
	
	choose $\gamma_1$ and $\gamma_2$,
	
	calculate $g_{i}(t) = \partial_{\hat{\mu}_{i}^{(0)}(t)}{l}(\tilde{\mu}_{i}(t), \hat{\mu}_{i}^{(0)}(t))$ and $h_{i}(t) = \partial^2_{\hat{\mu}_{i}^{(0)}(t)}{l}(\tilde{\mu}_{i}(t), \hat{\mu}_{i}^{(0)}(t))$.
	
	\vspace{0.2cm}
	\For{$k=1,...,K$}{
		
		\For{$d=1,...,D^{(k)}$}{
			\vspace{0.2cm}
			\If{d is not a terminal node}{
				construct $I_d$
				
				calculate $g_{\cdot}(t) = \sum_{i\in I_d}g_{i}(t)$ and $h_{\cdot}(t) = \sum_{i\in I_d}h_{i}(t)$.
				
				\vspace{0.2cm}
				\For{$i=1,...,p$}{
					
					$\mathrm{gain} \leftarrow 0$ 
					
					\For{$j=1,...,m$}{
						
						split the node by $x_{i,j}$
						
						calculate $G_1$ from (\ref{eq:gain})
						
						$\mathrm{gain} \leftarrow \max(\mathrm{gain}, G_1)$
						
					}	
					
					\If{	$\mathrm{gain} \leq 0$ }{
						node $d$ is a terminal node
					}
				}

			}
			\vspace{0.2cm}
			update the tree topology
			
			update $\hat{\mu}^{(k)}(t) = \hat{\mu}^{(k-1)}(t)+T^{(k)}(\bm{x})$
			
			update $g_{i}(t) = \partial_{\hat{\mu}_{i}^{(k)}(t)}{l}(\tilde{\mu}_{i}(t), \hat{\mu}_{i}^{(k)}(t))$ and $h_{i}(t) = \partial^2_{\hat{\mu}_{i}^{(k)}(t)}{l}(\tilde{\mu}_{i}(t), \hat{\mu}_{i}^{(k)}(t))$.
			
		}
	}
	\caption{the Boost-R algorithm}
\end{algorithm}

\SingleSpacedXII
\vspace{8pt}
\subsection{Boost-R with Dynamic Features} \label{sec:dynamic}
Next, we extend Boost-R to handle both static and dynamic features. As discussed in Section \ref{sec:problem}, 
with both static and dynamic features, the recurrent event data can be denoted by 
$\mathfrak{D}=(\bm{y}_i,\bm{x}_i, \bm{z}_i(t))$, where $|\mathfrak{D}|=n$, $\bm{x}_i\in \mathbb{R}^p$, $\bm{y}_i\in \mathbb{R}^{r_i+1}$ and $\bm{z}_i(t):[0,\infty)\rightarrow \mathbb{R}^q$.

Incorporating dynamic features into a tree-based method is challenging especially when dynamic features have cumulative effects on the event process \citep{Bacchetti1995, BouHamad2009}. In other words, for individual $i$, its cumulative intensity depends on the entire history of the dynamic feature information, $\bm{z}_i^h(t)=\{\bm{z}_i(\tau); 0 \leq \tau \leq t\}$, associated with that individual. 
One approach, proposed in \cite{Liu2019}, is to split a tree node based on static features, while the data on each node are modeled by a separate model that explains the effects of dynamic features.
In other words, dynamic features are nested with static features. At each tree leaf, different individuals are associated with different time-dependent functions. 
This approach can be justified when sub-populations are mainly characterized by static attributes, while dynamic features are used for explaining the between-individual variation within a tree node (i.e., the variation between event processes for individuals sharing similar attributes).

Following the idea above, an additive-tree-based model with both static and dynamic features can be expressed as:
\begin{equation} \label{eq:additive_tree_2} 
\hat{\mu}^{(K)}(t;\bm{x}_i,\bm{z}_i^h(t)) = \sum_{k=1}^{K}T^{(k)}(\bm{x}_i,\bm{z}_i^h(t)) = \sum_{k=1}^{K}\sum_{d=1}^{D^{(k)}}f_d^{(k)}(t;\bm{z}_i^h(t)) I(\bm{x}_i \in R_d^{(k)})
\end{equation}
where $\hat{\mu}^{(K)}(t;\bm{x}_i,\bm{z}_i^h(t))$ is the ensemble estimator of the cumulative intensity for individual $i$ based on $K$ trees; and $T^{(k)}(\bm{x}_i,\bm{z}_i^h(t)) = \sum_{d=1}^{D^{(k)}}f_d^{(k)}(t;\bm{z}_i^h(t)) I(\bm{x}_i \in R_d^{(k)})$ with $D^{(k)}$ denoting the number of leaves of tree $k$ and $f_d^{(k)}(t;\bm{z}_i^h(t))$ being a time-dependent function depending on $\bm{z}_i^h(t)$. It is easy to see how the idea of \cite{Liu2019} is embedded into (\ref{eq:additive_tree_2}): a tree node is split based on static features, and the data on each node are modeled by a second model that incorporates the dynamic features. 

The same idea behind Algorithm 1 can be adopted and modified to construct the boosting trees (\ref{eq:additive_tree_2}) with dynamic features. However, before we present the extended algorithm, some necessary modifications are needed:
\begin{itemize}
	\item Because it is less realistic to assume a parametric form for $f_d^{(k)}(t;\bm{z}_i^h(t))$ in (\ref{eq:additive_tree_2}), we adopt the non-parametric approach and model $f_d^{(k)}(t;\bm{z}_i^h(t))$ as
	\begin{equation} \label{eq:f} 
	\begin{split}
	f_d^{(k)}(t;\bm{z}_{i}^h(t)) & = \sum_{l}^{q}\int_{0}^{t}b_l(z_{i,l}(\tau);\bm{\beta}_l^{(d)})d\tau \\
	& =\sum_{l}^{q}\sum_{j}^{u+v}\left(\beta_{j,l}^{(d)}\int_{0}^{t}B_{j,v}(z_{i,l}(\tau))d\tau\right)
	\end{split}
	\end{equation}
	where $b_l$ is a linear combination of B-splines bases, $\bm{\beta}_l$ are the control or de Boor points, $B_{j,v}$ is the $j$th B-splines basis function of order $v$, and $u$ is the number of internal knots. 
	
	\item The regularization term (\ref{eq:penalty}) needs to be modified. Note that, for Boost-R with only static features, all individuals on a node $d$ share the same function $f_d(t)$. However, when dynamic features are included, individuals on the same node (i.e., individuals share the same static feature $\bm{x}$) are associated with different $f_d(t;\bm{z}^h(t))$ because these individuals typically possess different history of dynamic feature information. Hence, it is no longer meaningful to use the shrinkage strategy by penalizing $||f_{d}(t;\bm{z}^h(t))||_2^2$, which depends on dynamic features. This consideration motivates us to adopt the idea of Group Lasso, and obtain a set of $K$ trees by minimizing
	\begin{equation} \label{eq:objective2}
	\mathrm{minimize} \quad \mathfrak{L}^{(K)} = \sum_{i=1}^{n}l(\tilde{\mu}_i(t), \hat{\mu}^{(K)}(t;\bm{x}_i,\bm{z}^h_i(t))) + \gamma_1\sum_{k=1}^{K} D^{(k)} + \frac{1}{2} \gamma_2 \sum_{d=1}^{D^{(k)}} \sum_{l}^{q} ||\bm{\beta}_{\cdot,l}^{(d)}||_2.
	\end{equation}
	where $\bm{\beta}_{\cdot,l}^{(d)} = (\beta_{1,l}^{(d)},\beta_{2,l}^{(d)},\cdots, \beta_{u+v,l}^{(d)})$ for $l=1,2,\cdots,q$.
	The last term in (\ref{eq:objective2}) borrows the idea from Group Lasso, which has been widely used for model selection with grouped variables \citep{Yuan2007}. In (\ref{eq:objective2}), $||\bm{\beta}_{\cdot,l}^{(d)}||_2$ measures the size of $\bm{\beta}_{\cdot,l}$ which corresponds to dynamic feature $l$, i.e., the $l$th group in Group Lasso. If the $l$th feature turns out to be less important, it is necessary to drop the entire group vector, $\bm{\beta}_{\cdot,l}$, on a tree node.
	
	\item The node splitting procedure needs to be modified. In Section \ref{sec:Boost-R}, (\ref{eq:GB4}) and (\ref{eq:gain}) are obtained from (\ref{eq:GB3}) because all individuals on node $d$ (i.e. for all $i\in I_d$) share the same function $f_d(t)$. For the same reason, $g_{\cdot}(t) = \sum_{i\in I_d}g_{i}(t)$ and $h_{\cdot}(t) = \sum_{i\in I_d}h_{i}(t)$ can be properly defined. When dynamic features are included, individuals on the same node are not associated with the same $f_d(t;\bm{z}^h(t))$ as individuals are associated with different dynamic features. As a result, $g_{\cdot}(t)$ and $h_{\cdot}(t)$ cannot be defined and the ``gain'' in (\ref{eq:gain}) needs to be modified as:
	\begin{equation} \label{eq:gain2}
	G_2 = F_2(I_d)-F_2(I_d^{(L)})-F_2(I_d^{(R)})-2\gamma_1
	\end{equation}
	where 
	\begin{equation} 
	F_2(I) = \sum_{i\in I} \left\{\int_{0}^{c_i}g_i(t)f_d(t;\bm{z}^h_i(t))+\frac{1}{2}h_i(t)f_d^2(t;\bm{z}^h_i(t))dt  \right\} .
	\end{equation}
	
\end{itemize}

Based on the discussions above, the extended Boost-R algorithm with both static and dynamic features is summarized in Algorithm 2. 

\vspace{16pt}
\SingleSpacedXII
\begin{algorithm}[H]
	\SetAlgoLined
	\KwData{ ($\bm{x}_i,\bm{y}_i, \bm{z}_i(t)$) for $i=1,2,...,n$}
	
	\vspace{0.2cm}
	For all $i=1,...,n$:
	
	initialize $\mu^{(0)}(t)=0$,
	
	choose $\gamma_1$ and $\gamma_2$,
	
	calculate $g_{i}(t) = \partial_{\hat{\mu}_{i}^{(0)}(t)}{l}(\tilde{\mu}_{i}(t), \hat{\mu}_{i}^{(0)}(t))$ and $h_{i}(t) = \partial^2_{\hat{\mu}_{i}^{(0)}(t)}{l}(\tilde{\mu}_{i}(t), \hat{\mu}_{i}^{(0)}(t))$.
	
	\vspace{0.2cm}
	\For{$k=1,...,K$}{
		
		\For{$d=1,...,D^{(k)}$}{
			\vspace{0.2cm}
			\If{d is not a terminal node}{
				construct $I_d$
				
				
				\vspace{0.2cm}
				\For{$i=1,...,p$}{
					
					$\mathrm{gain} \leftarrow 0$ 
					
					\For{$j=1,...,m$}{
						
						split the node by $x_{i,j}$
						
						calculate $G_2$ from (\ref{eq:gain2})
						
						$\mathrm{gain} \leftarrow \max(\mathrm{gain}, G_2)$
						
					}	
					
					\If{	$\mathrm{gain} \leq 0$ }{
						node $d$ is a terminal node
					}
				}

			}
			\vspace{0.2cm}
			update the tree topology
			
			update $\hat{\mu}^{(k)}(t) = \hat{\mu}^{(k-1)}(t)+ T^{(k)}(\bm{x})$
			
			update $g_{i}(t) = \partial_{\hat{\mu}_{i}^{(k)}(t)}{l}(\tilde{\mu}_{i}(t), \hat{\mu}_{i}^{(k)}(t))$ and $h_{i}(t) = \partial^2_{\hat{\mu}_{i}^{(k)}(t)}{l}(\tilde{\mu}_{i}(t), \hat{\mu}_{i}^{(k)}(t))$.
			
		}
	}
	\vspace{8pt}
	\caption{Boost-R algorithm with both static and dynamic features}
\end{algorithm}

\clearpage
\SingleSpacedXII
\raggedbottom

\section{Numerical Examples, R Code and Applications} \label{sec:numerical}
Numerical studies, including a case study, are presented in this section to generate some critical insights on how Boost-R performs and illustrate the applications of Boost-R. 

\vspace{8pt}
\subsection{Computer Code}
Boost-R has been implemented in R and leverages the parallel computing capabilities of R. The code is available at GitHub (\url{https://github.com/dnncode/Boost-R}), and the use of the R code is demonstrated throughout this section.

\vspace{8pt}
\subsection{Investigate the basic properties using \texttt{DATASET A} and \texttt{DATASET B}}
To develop some basic understanding of how Boost-R performs, we start with a simple numerical example involving only 200 individuals. For each individual, two static features are respectively sampled from the unit interval $[0,1]$. Let $x_{i,j}$ denote the value of feature $j$ associated with individual $i$ ($i=1,2,...,200$, $j=1,2$), the recurrent events of individual $i$ are simulated from a homogeneous Poisson process with the following intensity: 
\begin{equation}
\lambda_i = \begin{cases}
0.01 \quad \textrm{if }0 \leq x_{i,1}, x_{i,2}\leq 0.5 \\
0.10 \quad \textrm{if }0.5<x_{i,1}, x_{i,2}\leq 1\\
0.05 \quad \textrm{otherwise}
\end{cases}
\label{eq:lambda_example_1}
\end{equation} 
This data set is referred to as \textbf{\texttt{DATASET A}} in this paper. 

To illustrative the Boost-R algorithm, we start with some arbitrarily chosen values: $K=50$, $\gamma_1=300$ and $\gamma_2=100$. In the R code, the boosting trees are grown using the function \texttt{BoostR}: 
\begin{equation} \label{eq:BoostR}
\begin{split}
& \texttt{BoostR.out =  BoostR(data, X, K.value=50,} \\
& \quad\quad\quad\quad\quad\quad \texttt{gamma1.value=300, gamma2.value=100, D.max=4)}
\end{split}
\end{equation} 
where \texttt{data} contains the recurrent event data, \texttt{X} is a matrix that contains feature information, \texttt{K.value}, \texttt{gamma1.value} and \texttt{gamma2.value} are the specified values for $K$, $\gamma_1$ and $\gamma_2$, and \texttt{D.max} triggers the termination of the tree growing process once the number of leaves of a tree is not smaller than \texttt{D.max}. 
The function \texttt{BoostR} returns an object \texttt{BoostR.out}. Four R functions have been created to visualize the output of Boost-R. These functions include  \texttt{Plot\_Partition}, \texttt{Plot\_Individual}, \texttt{Plot\_Leaf}, and \texttt{Plot\_Imp}, which will be discussed next.

\begin{figure}[h!]
	\begin{center}
		\includegraphics[width=0.9\textwidth]{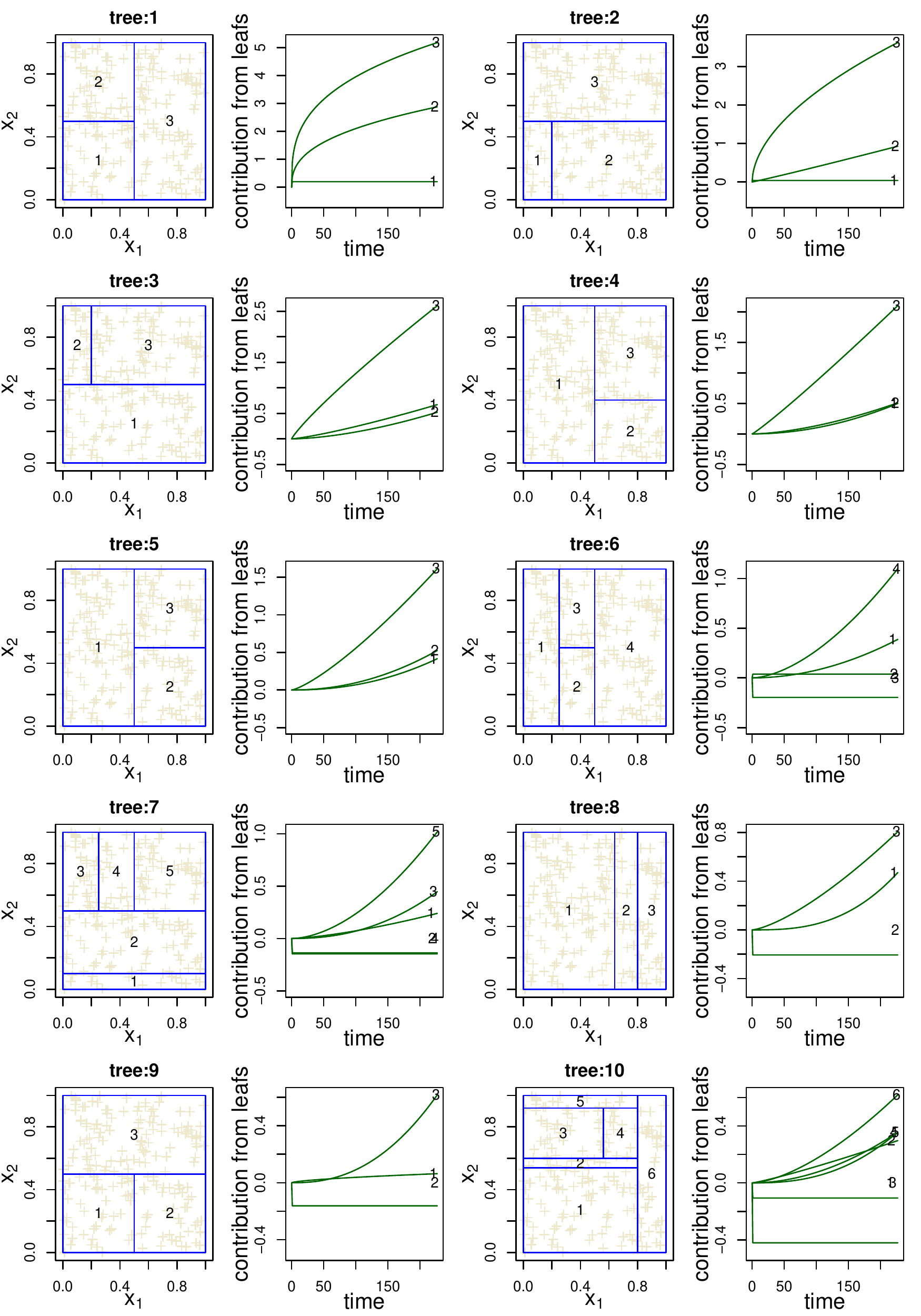}
	\end{center}
	\vspace{-0.1in}
	\caption{Columns 1 and 3 show the binary partitions of the feature space $[0,1]^2$ by the $K$ ($=10$) trees from the ensemble. Columns 2 and 4 show the contribution (i.e., $f_d^{(\cdot)}(t)$) to the estimated cumulative intensity function from each tree leaf of a tree.}
	\label{fig:mcf_tree_example1} 
	\vspace{-0.1in}
\end{figure}

The function, \texttt{Plot\_Partition(BoostR.out)}, visualizes the binary partitions by individual trees, as well as $f_d^{(\cdot)}(t)$ on each partition. 
Figure \ref{fig:mcf_tree_example1} shows the first 10 trees obtained from the Boost-R algorithm. Columns 1 and 3 of this figure present the binary partitions of the feature space $[0,1]^2$ by individual trees. Columns 2 and 4 show the contribution (i.e., $f_d^{(\cdot)}(t)$) to the estimated cumulative intensity function from each tree leaf.

\clearpage
\clearpage

\begin{figure}[h!]
	\begin{center}
		\includegraphics[width=0.75\textwidth]{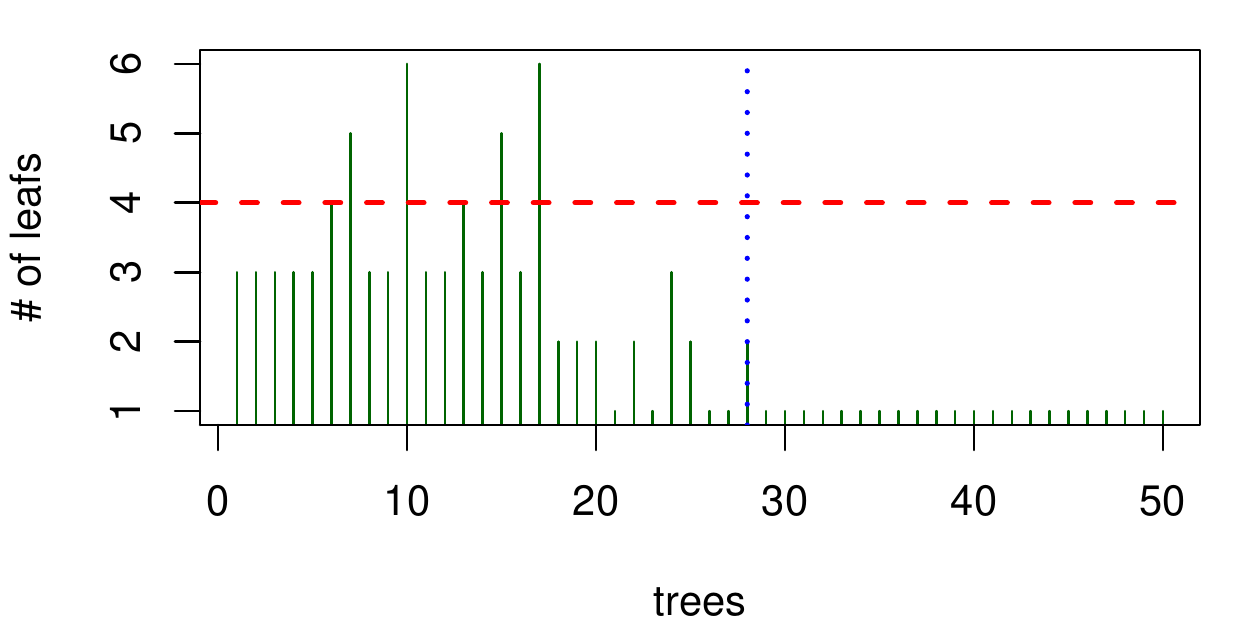}
	\end{center}
	\caption{Number of leaf nodes for the 50 trees. Here, we purposely overfit the data in order to generate some critical insights on how Boost-R performs.}
	\label{fig:n_leaf_example1} 
\end{figure}

It is seen from Figure \ref{fig:mcf_tree_example1} that each tree performs a binary partition of the feature space, which divides the 200 individuals into several sub-populations represented by tree leaves. For each sub-population, the contribution to the cumulative intensity function $f_d^{(\cdot)}(t)$ is computed. Three critical observations are obtained:
\begin{itemize}
	\item The idea behind boosting suggests that a new tree is added to explain only a small amount of variation not captured by previous trees. Hence, the newly added tree is used to perform some necessary adjustments to the output generated from previous trees (i.e., performance boosting). By examining the scale of the vertical axis in columns 2 and 4, it is easy to see that the amount of adjustment by a newly added tree becomes smaller as more trees have already been included in the ensemble. Using \texttt{Plot\_Individual(BoostR.out)}, Figure \ref{fig:converge_example1} shows how the final ensemble estimates for individuals 1 and 2 are obtained by aggregating outputs from individual trees.
	\begin{figure}[h!]
		\begin{center}
			\includegraphics[width=0.95\textwidth]{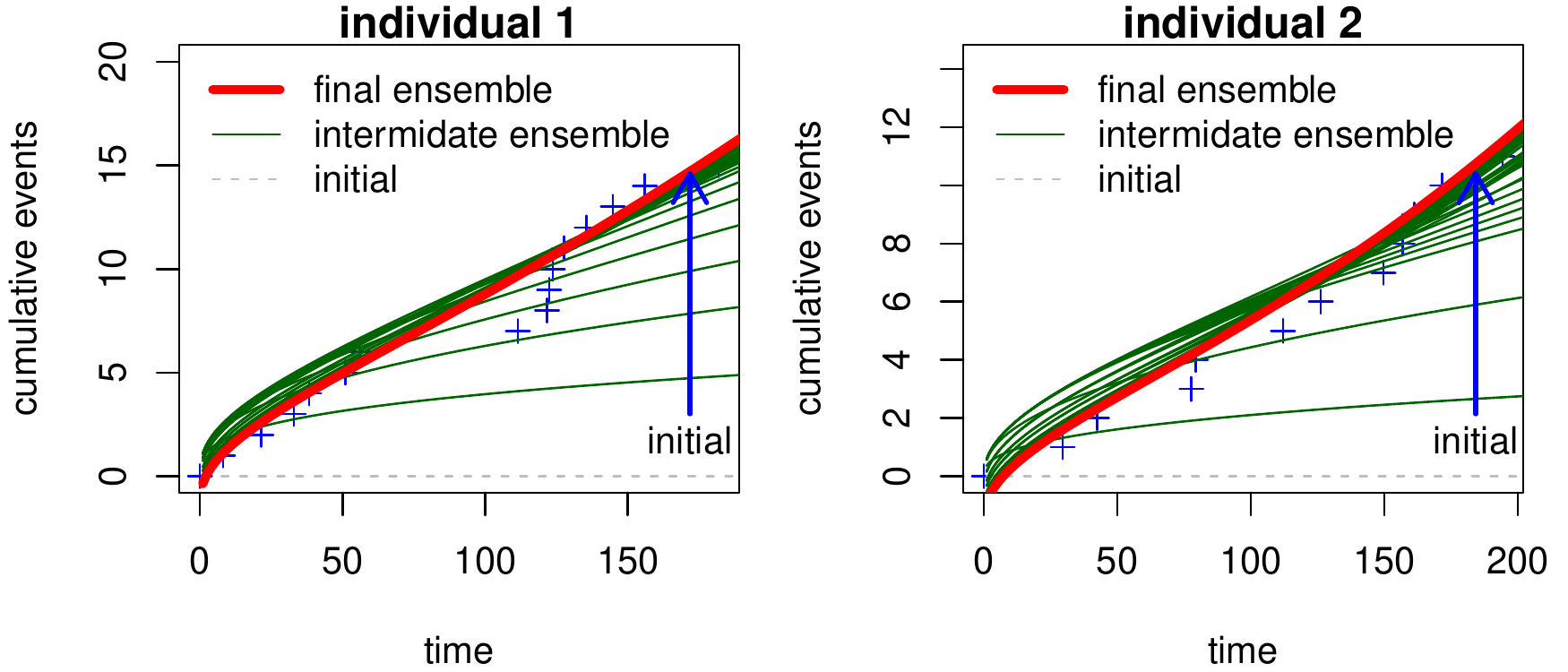}
		\end{center}
		\caption{The ensemble estimate of cumulative intensity functions by adding the contributions from individual boosting trees. The blue arrow shows how the initial cumulative intensity function (the horizontal dash line) converges to the ensemble estimate (the thick red curve) as more trees are grown.}
		\label{fig:converge_example1} 
		\vspace{-0.1in}
	\end{figure} 

	\item The amount of adjustment made by a newly added tree is not necessarily positive; for example, on leaves 2 and 4 of tree \#7. Typically, this happens when the estimated cumulative intensity is getting closer to the true function. Under such a circumstance, a newly added tree may suggest either increase or decrease the estimated cumulative intensity for certain sub-populations. 
	\item The key idea behind boosting trees is that each individual tree must be kept simple to form weak learners. In Boost-R, two mechanisms are used to regularize the tree complexity, i.e., the regularization term (\ref{eq:penalty}) and \texttt{D.max} in (\ref{eq:BoostR})
	specified by users. Figure \ref{fig:n_leaf_example1}, generated by \texttt{Plot\_Leaf(BoostR.out)}, shows the number of leaf nodes for the 50 trees. It is interesting to see that: (i)  
	Since the maximum number of tree leaves, \texttt{D.max}, is set to 4 in this example, the tree growing process is forced to stop once the number of leaves has exceeded 4 (note that, the final number of leaves is not necessarily 4 in this case if two or more nodes are split simultaneously before the algorithm ends). As shown in Figure \ref{fig:n_leaf_example1}, this rule applies to tree \#7, \#10, \#15 and \#17; (ii) For all other trees, the tree growing processes are terminated before the number of leaves has reached \texttt{D.max}. This observation justifies the effectiveness of the regularization (\ref{eq:penalty}) on tree complexity; (iii) After tree \#28, the remaining trees consist of only the root node. In this example, as most of the variation has been effectively explained by the first 28 trees, the gain of adding a new tree to the ensemble is outweighed by the penalty incurred by adding that new tree. Note that, it would not be possible to observe such a phenomenon if the regularization term (\ref{eq:penalty}) was removed (as there would be no penalty associated with increasing tree complexity). From this perspective, Figure \ref{fig:n_leaf_example1} also provides some insights on the choice of $K$, which will be investigated later. 
\end{itemize}

\begin{figure}[h!]
	\begin{center}
		\includegraphics[width=1\textwidth]{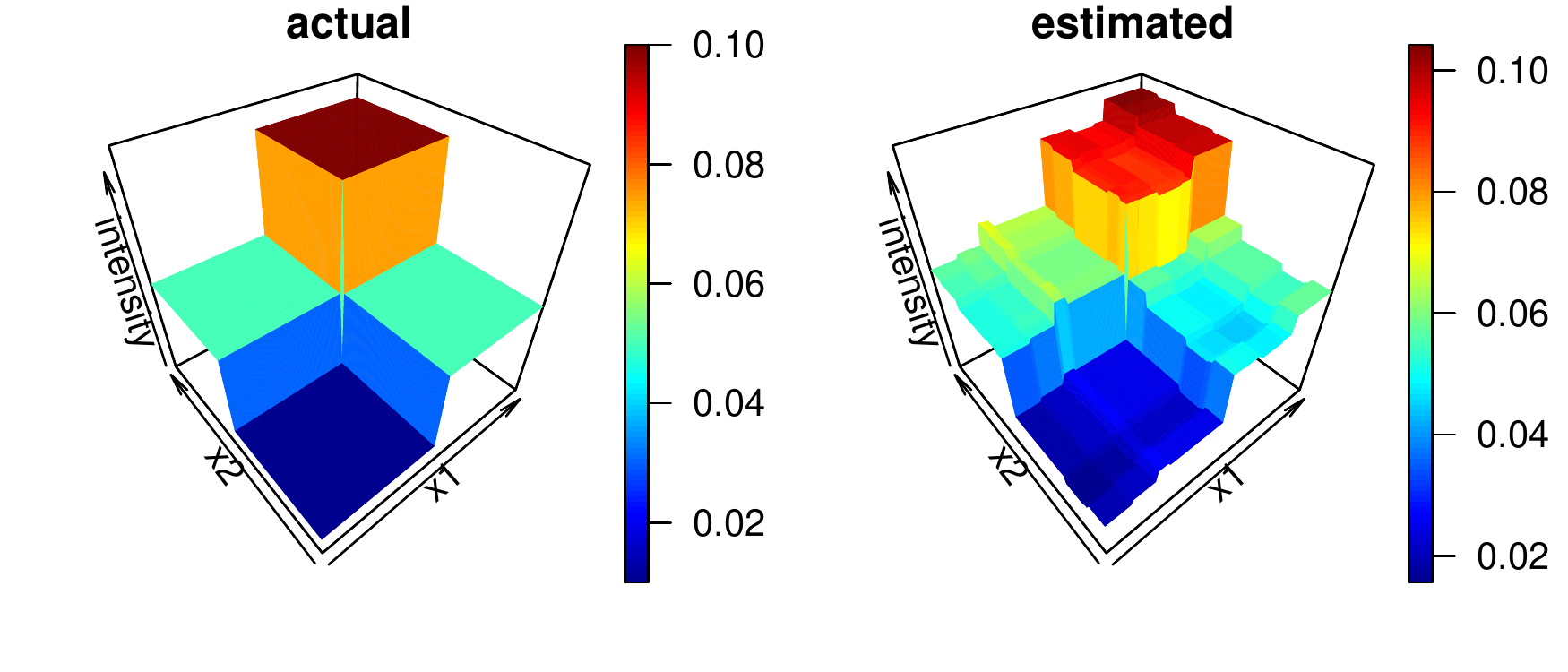}
	\end{center}
	\vspace{-0.2in}
	\caption{Learning the relationship between the cumulative intensity function and features. Left panel: the true relationship; Right panel: the estimated relationship by Boost-R.}
	\label{fig:space_example1} 
	\vspace{-0.1in}
\end{figure}

The boosting trees obtained from Boost-R are capable of accurately capturing the interactions between the recurrent event process and features. Figure \ref{fig:space_example1} shows both the actual (left panel) and estimated (right panel) relationship between the cumulative intensity function and features. Note that, since the Boost-R algorithm does not assume that the intensity function is time-invariant, the intensity shown on the right panel is the average intensity over time over the feature space. Figure \ref{fig:space_example1} clearly demonstrates the potential of Boost-R in capturing the relationship between recurrent event processes and features. More complex scenarios are investigated in Section \ref{sec:3.2}. 

\vspace{6pt}
Boost-R requires one to specify the parameters, $\gamma_1$, $\gamma_2$ and $K$. The first two parameters $\gamma_1$ and $\gamma_2$ control the complexity of individual trees, while $K$ determines the number of trees in the ensemble. In general, $K$ needs to be sufficiently large so that there will be enough number of trees, but not too large which causes overfitting. 

\begin{figure}[h!]
	\begin{center}
		\includegraphics[width=1\textwidth]{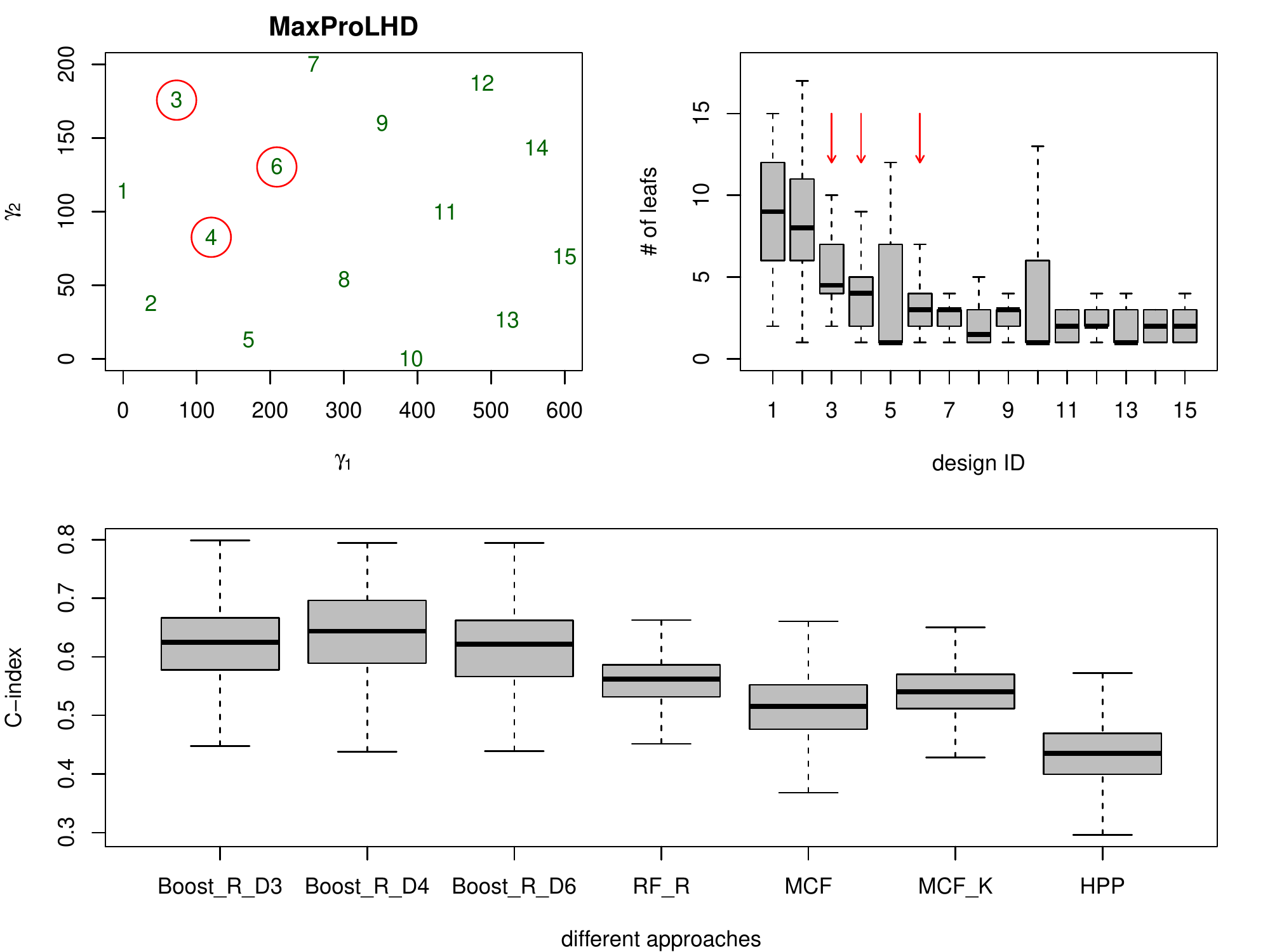}
	\end{center}
	\caption{Model improvements and comparison. Top left panel: the MaxProLHD design for 15 combinations of $\gamma_1$ and $\gamma_2$; Top right panel: the number of tree leaves for trees in an ensemble; Bottom: comparison of C-index for different models}
	\label{fig:tuning_example1} 
	\vspace{-0.1in}
\end{figure}

As a common strategy in practice, we explore the suitable values for $\gamma_1$ and $\gamma_2$, leaving $K$ as the primary parameter \citep{Hastie2009}. Although it is theoretically possible to perform a grid search for the best combinations of $\gamma_1$ and $\gamma_2$ on a two-dimensional space, such an approach may not be practical nor necessary in practice when it is computationally intensive to run Boost-R on big datasets. Hence, we resort to a powerful tool in computer experiments---the space-filling designs \citep{Joseph2016}. The idea of space-filling designs is to have points everywhere in the experimental region with as few gaps as possible, which serves our purpose very well. The top left panel of Figure \ref{fig:tuning_example1} shows the Maximum Projection Latin Hypercube Design (MaxProLHD, \cite{Joseph2015}) of 15 runs with different combinations of $\gamma_1$ and $\gamma_2$, where the experimental ranges for these two parameters are respectively $[0,600]$ and $[0,200]$. 
The top right panel of Figure \ref{fig:tuning_example1} shows the box plot of the number of tree leaves per tree in an ensemble, for each combination of $\gamma_1$ and $\gamma_2$. Since the key idea behind boosting trees is that each individual tree needs to be kept simple with 4 to 8 leaves \citep{Hastie2009}, we quickly identify that Designs \#3, \#4 and \#6 provide the most suitable combinations of $\gamma_1$ and $\gamma_2$. From the top left panel of Figure \ref{fig:tuning_example1}, these three design points are adjacent to each other, indicating that the appropriate choices for $\gamma_1$ and $\gamma_2$ are respectively within $[75,220]$ and $[75,175]$. If necessary, a more refined search can be perform in a much smaller experimental region. 

\clearpage

We compare the performance of Boost-R with that of existing approaches, including 1) \textit{RF-R}: Random Forest for Recurrence Data; 2) \textit{MCF}: the nonparametric estimation for Mean Cumulative Function (MCF) without using feature information; 3) \textit{MCF-K}: the nonparametric estimation for MCF utilizing only the data from the $K$ nearest neighbors of an individual (the distance between two individuals is defined by the Euclidean distance in the feature space); and 4) \textit{HPP}: the HPP model with a log-linear intensity of static features.

Cross-validation is used to evaluate the performance of all candidate approaches. For each approach, the data set is randomly divided into a training set (150 individuals) and a testing set (50 individuals). The model is trained using the training set, and the prediction C-index is used as the performance measure. The above procedure is repeated for 500 times in order to generate the boxplot of C-indices at the bottom of Figure \ref{fig:tuning_example1}. It is seen that, the Boost-R (based on the three best designs, Designs \#3, \#4 and \#6), yields a higher C-index than other competing methods.  
Here, the C-index, or Harrell's concordance index, was firstly proposed in \cite{Harrell1982} for evaluating the amount of information a medical test provides about individual patients. For the problem considered in this paper, the C-index can be interpreted as the empirical probability of correctly ranking any two individuals in terms of their cumulative number of failures, and can be calculated in the following way: 1) form all pairs of individuals from the testing data set. 2) for each pair, rank the two individuals based on the cumulative number of events up to a given time. 3) for each pair, rank the two individuals based on the predicted cumulative number of events up to the same time. If the predicted rankings are consistent with the observed rankings, let $C_i=1$ for pair $i$, otherwise  $C_i=0$; and 4) the C-index for a testing data set is the empirical probability of correctly ranking any two individuals.

\begin{figure}[h!]
	\begin{center}
		\includegraphics[width=0.75\textwidth]{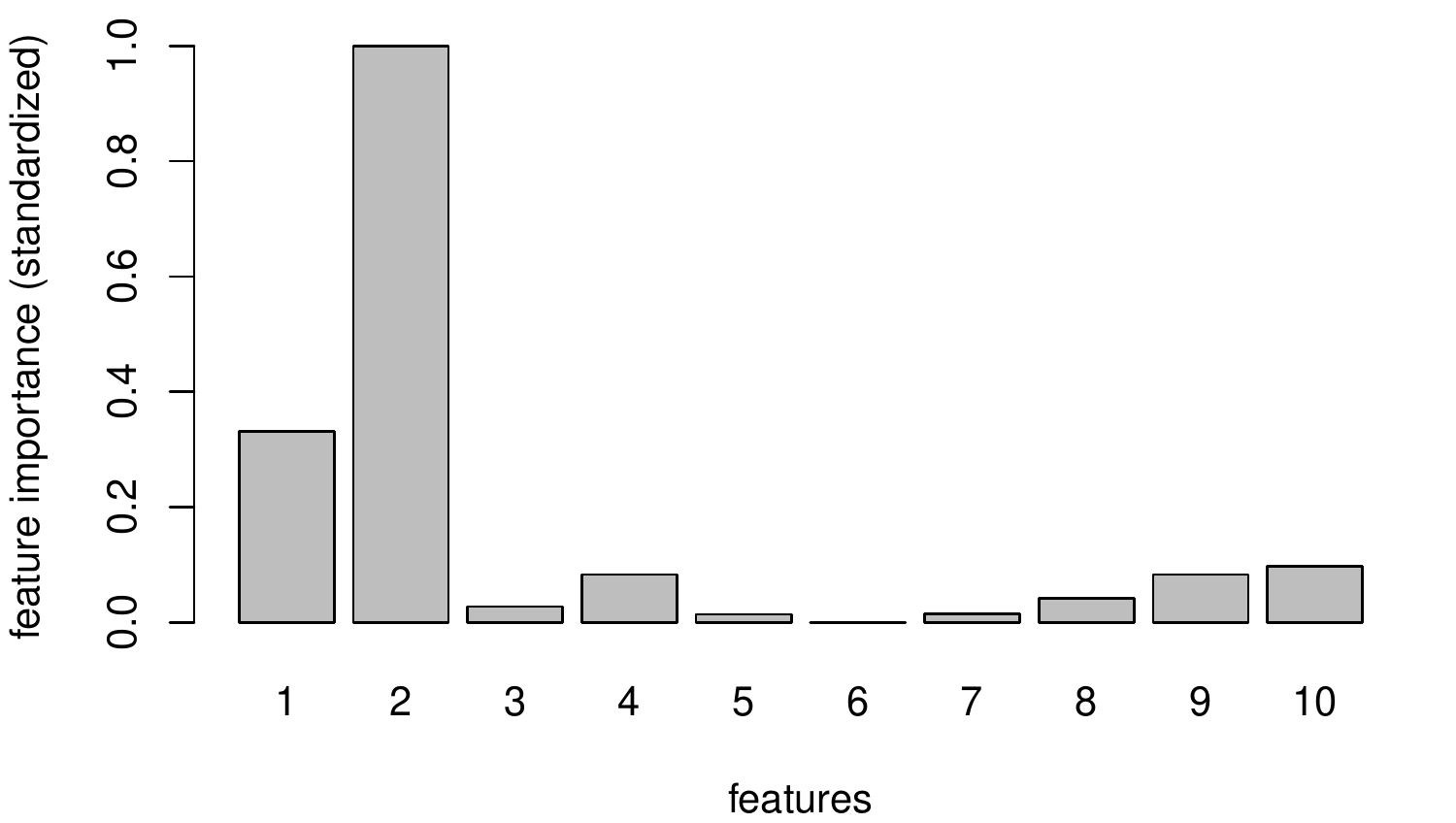}
	\end{center}
	\caption{Standardized feature importance measured by the total gain achieved by splitting tree nodes based on a given feature; see (\ref{eq:importance})}
	\label{fig:importance_example1} 
\end{figure}

Finally, to illustrate the feature selection capability of Boost-R, we include eight randomly generated redundant covariates, $\{x\}_{i=3}^{10}$, in \textbf{\texttt{DATASET A}}. The new data set with redundant covariates is referred to as \textbf{\texttt{DATASET B}}. We re-run Boost-R using \textbf{\texttt{DATASET B}}, and the function, \texttt{Plot\_Imp(BoostR.out, standardize=TRUE)}, shows the importance for the 10 features as defined in (\ref{eq:importance}); see Figure \ref{fig:importance_example1}. Here, we standardize the importance measure (\ref{eq:importance}) such that the highest and lowest importance are respectively 1 and 0. It is immediately seen that the algorithm successfully identify the correct features, $x_1$ and $x_2$. 

\subsection{More complicated scenarios using \texttt{DATASET C} and \texttt{DATASET D}} 
\label{sec:3.2}
In the previous illustrative example, \texttt{DATASETS A} and \texttt{B} are simulated from a simple HPP with intensity (\ref{eq:lambda_example_1}). To investigate the learning capabilities of Boost-R, we consider more complicated interactions between event processes and features. In particular, 
\texttt{DATASET C} and \texttt{DATASET D} respectively consist of the simulated recurrent event data from 1000 individuals. For both datasets, two features, $x_1$ and $x_2$, are sampled from the unit interval $[0,1]$. In addition, 

\begin{itemize}
	\item For \texttt{DATASET C}, the event times for individual $i$ are simulated from a non-homogeneous Poisson process using the thinning method \citep{Lewis1979} with the following intensity function:
	\begin{equation}
	\lambda_i(t) = \begin{cases}
	1.5t^{-0.5} \quad \textrm{if }(x_{i,1}-0.5)^2+(x_{i,2}-0.5)^2 \leq 0.04 \\
	t^{-0.5} \quad\quad \textrm{if }0.04<(x_{i,1}-0.5)^2+(x_{i,2}-0.5)^2 \leq 0.16\\
	0.5t^{-0.5} \quad \textrm{otherwise}
	\end{cases}
	\label{eq:lambda_dataC}
	\end{equation} 
	\item For \texttt{DATASET D}, the event times for individual $i$ are simulated from a non-homogeneous Poisson process with the following intensity function:
	\begin{equation}
	\lambda_i(t) = 0.01t^{0.5}\exp(0.5(x_{i,1}-0.5)^2+2(x_{i,2}-0.5)^2).
	\label{eq:lambda_dataD}
	\end{equation} 
\end{itemize}

\begin{figure}[h!]
	\begin{center}
		\includegraphics[width=0.85\textwidth]{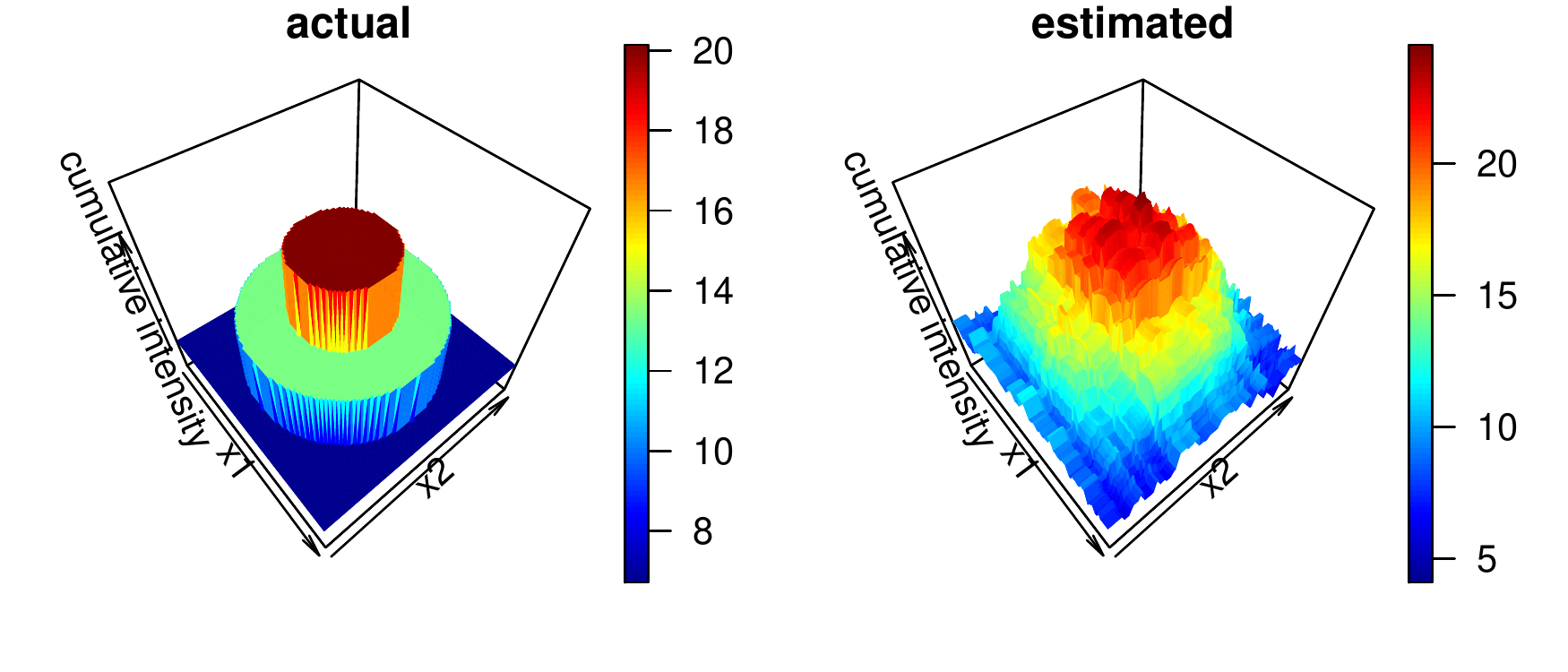}
	\end{center}
	\vspace{-0.2in}
	\caption{Actual (left) and estimated (right) cumulative intensity at time 50 based on \texttt{DATASET C}}
	\label{fig:space_dataC} 
	\vspace{-0.1in}
\end{figure}
\begin{figure}[h!]
	\begin{center}
		\includegraphics[width=0.85\textwidth]{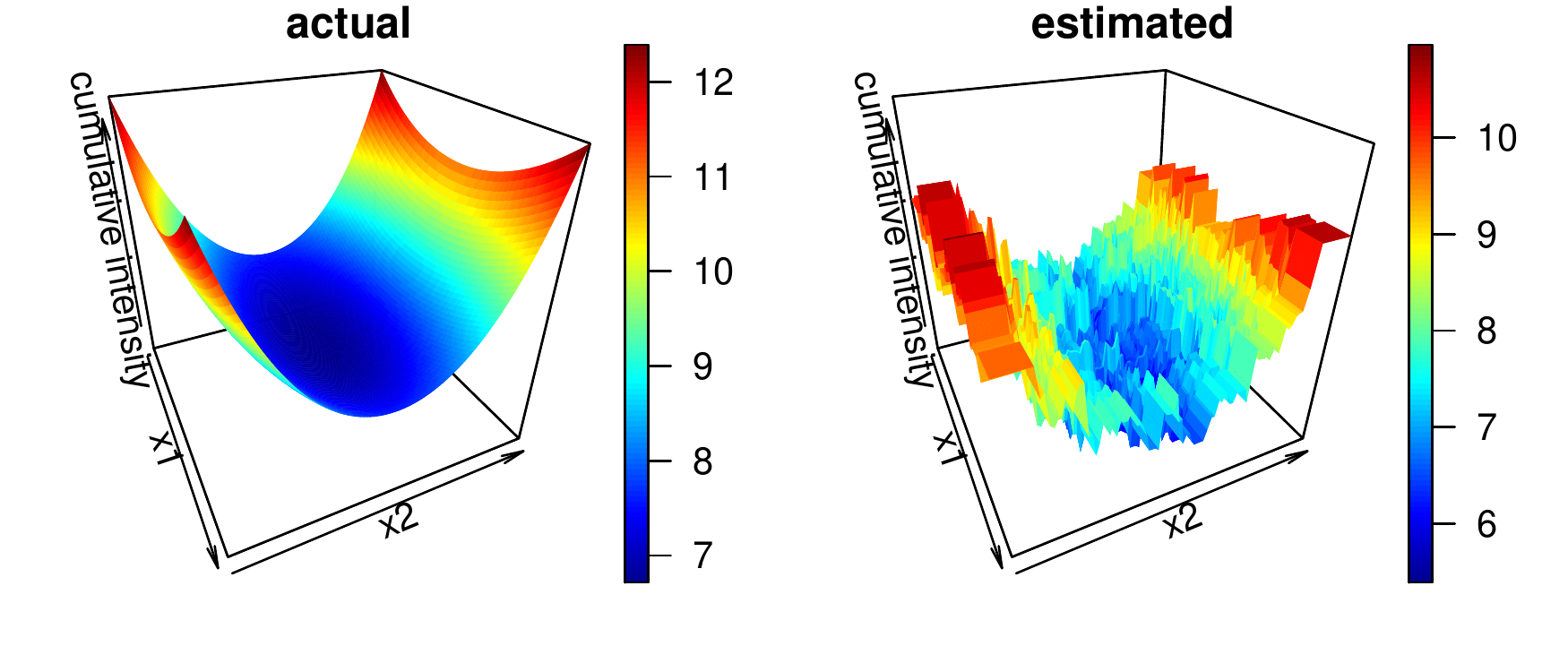}
	\end{center}
	\vspace{-0.2in}
	\caption{Actual (left) and estimated (right) cumulative intensity at time 100 based on \texttt{DATASET D}}
	\label{fig:space_dataD} 
	\vspace{-0.1in}
\end{figure}

We run the Boost-R algorithms using both data sets. For \texttt{DATASET C}, we let $K=500$, $\gamma_1=10$ and $\gamma_2=5$. For \texttt{DATASET D}, we let $K=300$, $\gamma_1=100$ and $\gamma_2=100$. 
Based on \texttt{DATASET C}, Figure \ref{fig:space_dataC} shows both the actual cumulative intensity (left panel) and estimated cumulative intensity (right panel) at time 50 over the feature space. Based on \texttt{DATASET D}, Figure \ref{fig:space_dataD} shows both the actual cumulative intensity (left panel) and estimated cumulative intensity (right panel) at time 100 over the feature space. It is seen that, Boost-R successfully captures the complex relationship between cumulative intensity and features, as specified in (\ref{eq:lambda_dataC}) and (\ref{eq:lambda_dataD}). Such complicated and highly nonlinear relationships can hardly be specified (unless they are known in advance) when traditional parametric approaches are used.


\vspace{8pt}
\subsection{Modeling the Failure Processes of Oil and Gas Wells} \label{sec:case}

To run Boost-R with both static and dynamic features (i.e., Algorithm 2), we model $f_d^{(k)}(t;\bm{z}_i(t))$ in (\ref{eq:f}) by cubic splines with two  internal knots, and let $K=300$, $\gamma_1=300$, $\gamma_2=100$, $v=3$ and $u=2$. 
In our R code, the boosting trees are grown using the function \texttt{BoostR2}: 
\begin{equation}
\begin{split}
& \texttt{BoostR.out = BoostR2(data, X, Z=z.list, K.value=300,} \\
& \quad\quad\quad\quad\quad \texttt{gamma1.value=300, gamma2.value=100, u.value=2, v.value=3, D.max=4)}
\end{split}
\end{equation} 
where \texttt{data} contains the recurrent event times, \texttt{X} is a matrix that contains the static feature information, $Z$ contains the dynamic feature information, \texttt{K.value}, \texttt{gamma1.value}, \texttt{gamma2.value}, \texttt{u.value} and \texttt{v.value} are respectively the specified values for $K$, $\gamma_1$, $\gamma_2$, $u$ and $v$, and the last input \texttt{D.max} determines the termination of the tree growing process once the number of leaves per tree reaches or exceeds \texttt{D.max}. 
The function \texttt{BoostR2} returns an objective that contains the output of Boost-R with both static and dynamic features, i.e., Algorithm 2. The output generated by \texttt{BoostR2} can be visualized by the following functions: \texttt{Plot\_Partition}, \texttt{Plot\_Individual}, \texttt{Plot\_Leaf}, \texttt{Plot\_Imp}, and \texttt{Plot\_Interaction}.

\begin{figure}[h!]
	\begin{center}
		\includegraphics[width=0.65\textwidth]{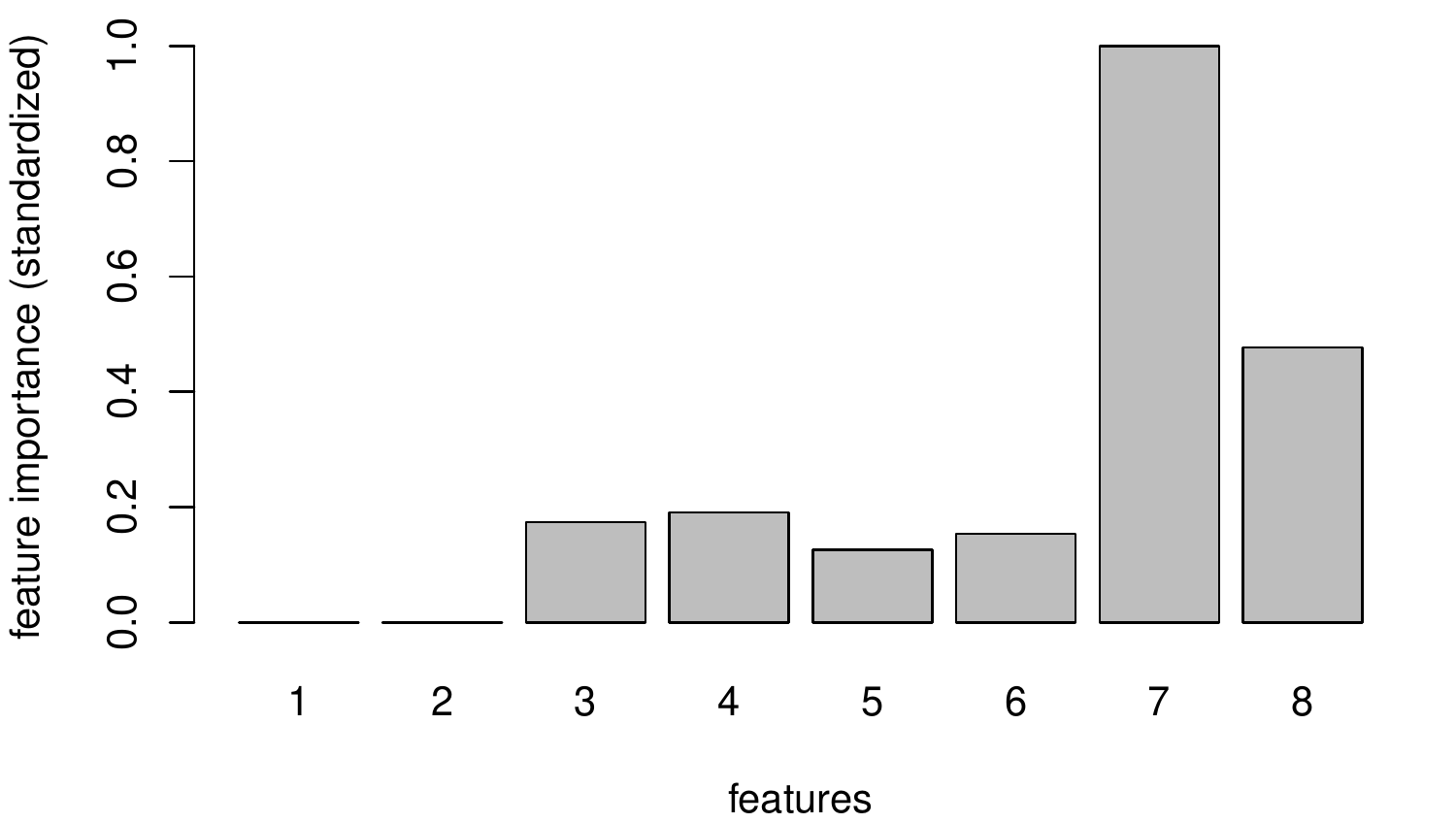}
	\end{center}
	\caption{Feature importance for the 8 static well attributes, which is measured by the total gain achieved by splitting tree nodes based on a given feature; see (\ref{eq:importance})}
	\label{fig:importance} 
	\vspace{-0.1in}
\end{figure}

Figure \ref{fig:importance} is generated by \texttt{Plot\_Imp(BoostR.out, standardize=TRUE)}, and shows the importance of the eight static system attributes as the total gain respectively achieved by splitting tree nodes based on each feature. The algorithm clearly identifies $x_7$ and $x_8$, the geo-locations, as the two most important system attributes. Because wells at similar geographical locations share common, but unknown, environmental conditions (e.g., temperature and humidity variation, soil type, contamination, etc.), geo-locations serve as important proxies in capturing those unknown environmental factors which may lead to some important spatial patterns such as trend and clustering. 
The results shown in Figure \ref{fig:importance} confirm that these unknown environmental factors significantly influence the system failure processes, leading to different failure patterns among these well systems.

Next, we re-run the Boost-R algorithm by retaining the two static features, $x_7$ and $x_8$, and the dynamic gearbox torque. Although one might as well include $x_3,x_4,...,x_6$, keeping only $x_7$ and $x_8$ allows us to effectively visualize the interesting interaction between the estimated B-splines coefficients $\bm{\beta}_{\cdot,1}$ and spatial locations $x_7$ and $x_8$. 
\begin{figure}[h!]
	\begin{center}
		\includegraphics[width=1\textwidth]{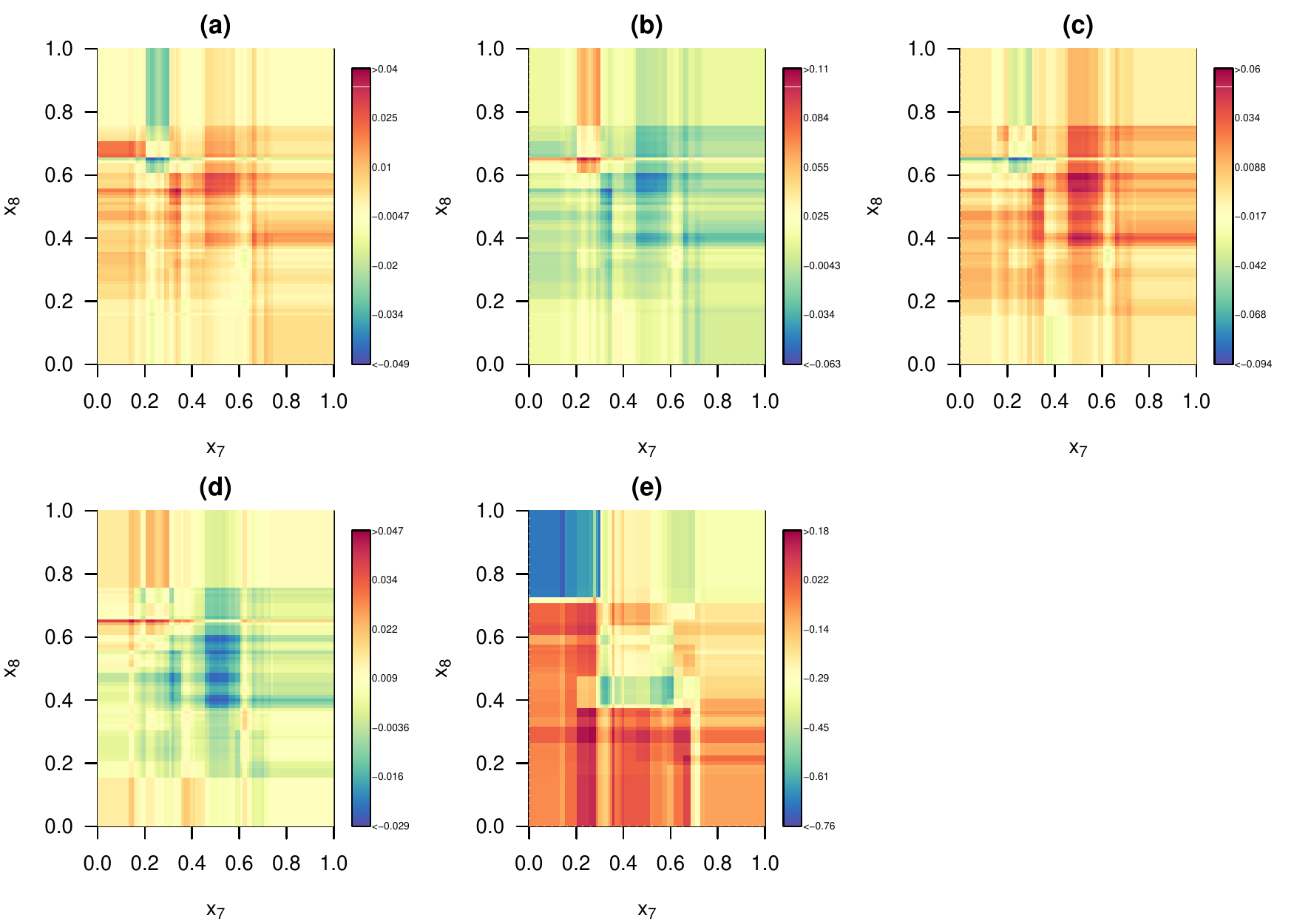}
	\end{center}
	\caption{Aggregated B-splines coefficients over the spatial domain. Subplots (a) to (e) respectively correspond to the coefficients ${\beta}_{1,1}$, ${\beta}_{2,1}$, $\cdots$, ${\beta}_{5,1}$.}
	\label{fig:interaction} 
	\vspace{-0.1in}
\end{figure}

Because cubic splines with two internal knots are used in this example, we have $\bm{\beta}_{\cdot,1} = ({\beta}_{1,1},{\beta}_{2,1},\cdots,{\beta}_{5,1})$. Figure \ref{fig:interaction}, which is generated by \texttt{Plot\_Interaction(BoostR.out)}, provides a spatially aggregated view of the five estimated B-splines coefficients over the spatial domain. 
Note that, each boosting tree partitions the spatial domain $[0,1]^2$ into several rectangular areas and the estimated value of $\bm{\beta}_{\cdot,1}$ is obtained for each area. 
Because $\bm{\beta}_{\cdot,1}$ can be viewed as the effects of the dynamic feature on the recurrent event processes, it is immediately seen that such effects vary over the spatial domain. In other words, the cumulative failure intensities at different geo-locations are influenced by the operational conditions. For example, the aggregated values of $\beta_{1,1}$ appears to be lower in the area where $0.2 < x_7 < 0.25$ and $x_8>0.6$, while the aggregated value of $\beta_{2,1}$ is larger in approximately the same area. This observation strongly demonstrates the effectiveness of Boost-R in capturing the interactions between static and dynamic features, by leveraging the advantages of binary tree structures. 

\begin{figure}[h!]
	\begin{center}
		\includegraphics[width=1\textwidth]{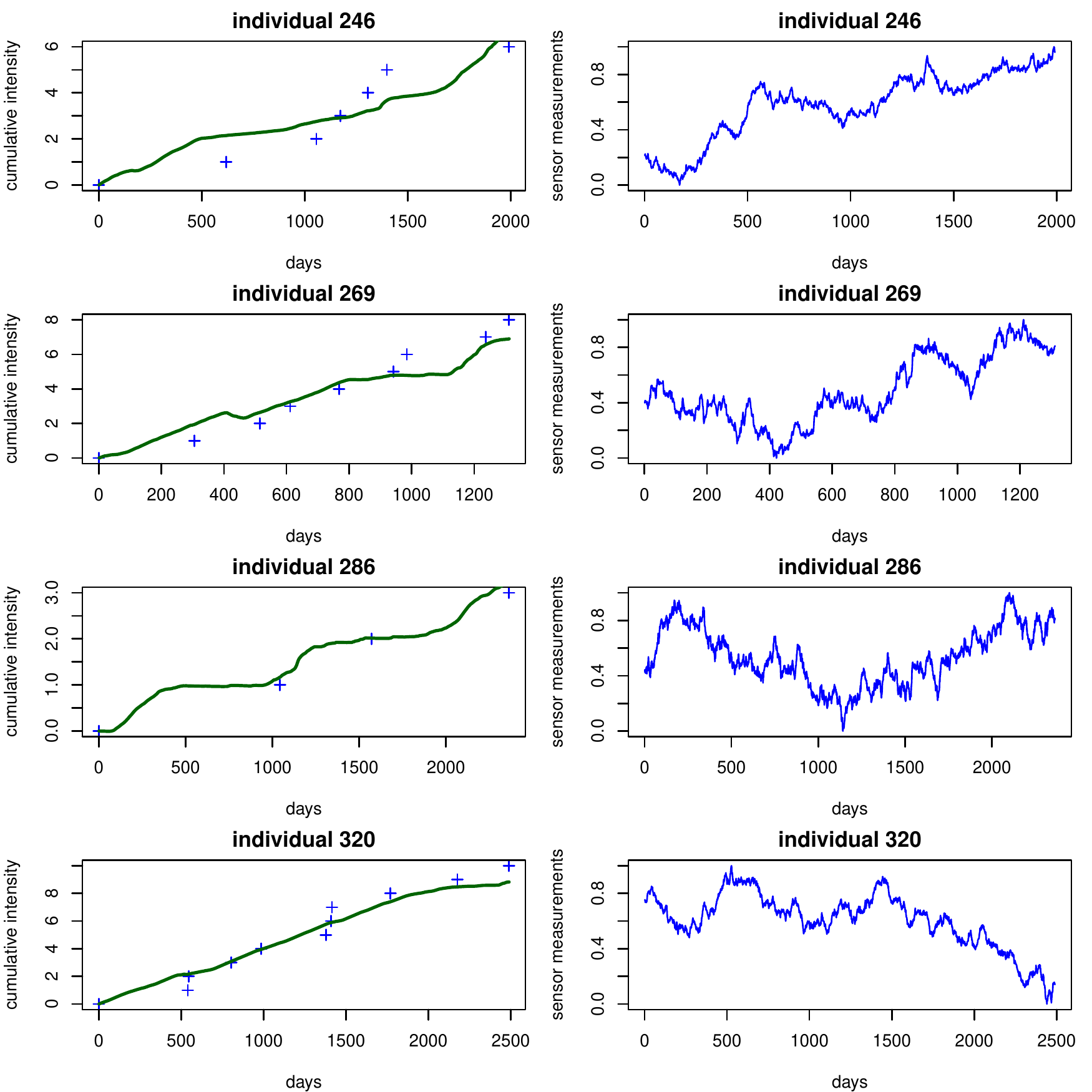}
	\end{center}
	\caption{Estimated cumulative intensity (left) and gearbox torque (right) for selected well systems.}
	\label{fig:estimate} 
\end{figure}

Using \texttt{Plot\_Individual(BoostR.out)}, Figure \ref{fig:estimate} shows the estimated cumulative failure intensity and cumulative failure counts of four selected well systems (left column). The observed gearbox torque for these well systems are also shown in the right column. We see that, Boost-R successfully estimates the cumulative failure intensity for heterogeneous individuals with diverse system attributes (static features) and operating conditions (dynamic feature). Such an observation is  encouraging and demonstrates the potential of Boost-R for recurrent event data analytics: the system heterogeneity is addressed by the ``divide-and-conquer'' structure of binary trees, and the non-parametric approaches (including the binary tree and B-splines) are used to capture the complex, often non-linear, interactions between recurrent event processes and feature information without imposing parametric assumptions. 

In the Appendices, we provide additional application examples and comparison studies between Boost-R and other methods. In particular, we also provide some discussions on the potential use of XGBoost for recurrent event data. 

\vspace{8pt}
\section{Conclusions} \label{sec:conclusions}
This paper proposed an additive-tree-based statistical learning approach, known as Boost-R (\textbf{Boost}ing for \textbf{R}ecurrence Data), for modeling recurrent event data with both static and dynamic feature information. The technical details behind Boost-R have been presented. Gradient boosting algorithms have been developed to obtain an ensemble of correlated trees that generate the estimated cumulative intensity functions characterizing the recurrent event processes given feature information. To our best knowledge, Boost-R is the first gradient boosted additive-tree-based model for recurrence data with both static and dynamic features.

The advantages of Boost-R are due to three salient features behind this approach: (i) Boost-R leverages the ``divide-and-conquer'' structure of binary trees to address the inevitable heterogeneity among a large population of individuals; (ii) the non-parametric nature of the algorithm (e.g., binary tree and B-splines) enables us to capture the complex and non-linear relationship between event processes and features, which may not be adequately captured by parametric approaches; (iii) Boost-R is built into the framework of gradient boosted trees, which has proven to be one of the most successful statistical learning approaches over the past decade.
Comprehensive numerical studies, including a case study, have been performed to demonstrate the advantages of Boost-R. R code has been made available on GitHub to facilitate the adoption of this new approach. 


\vspace{16pt}
\bibliographystyle{asa}
\bibliography{references}


%
%
%

\vspace{30pt}
 \begin{APPENDICES}
 	
 	\section{Discussions on the Ad-Hoc Use of XGBoost for Recurrence Data}
 	
 	[*Appendix A can be moved to supplementary materials if necessary] \\
 	 
 	If time $t$ is treated as an additional feature, XGBoost is sometimes used in industry to model recurrent event data. Although the off-the-shelf XGBoost can learn the relationship between the cumulative number of events and time and other features, such an \textit{ad hoc} use of XGBoost may have a few major limitations:
 	
 	$\triangleright$ When time is treated as another feature, the predicted cumulative intensity is typically ``bumpy'' because the cumulative events at different times are learned independently.  
 	To illustrate this point, Figure \ref{fig:bumpy} below shows how the predicted cumulative intensity from XGBoost and the proposed Boost-R typically look like. Because XGBoost models the number of cumulative failures at different times independently, the predicted intensity at a given time depends on the number of observations available at that time and the observed cumulative events from those available samples at that time. As a result, it is not surprising that the predicted cumulative intensity from XGBoost is rarely smooth. Boost-R, on the other hand, learns a smooth time-dependent function at each tree leaf, and the ensemble prediction of the cumulative intensity is also smooth.  
 	\begin{figure}[h!]
 		\begin{center}
 			\includegraphics[width=0.7\textwidth]{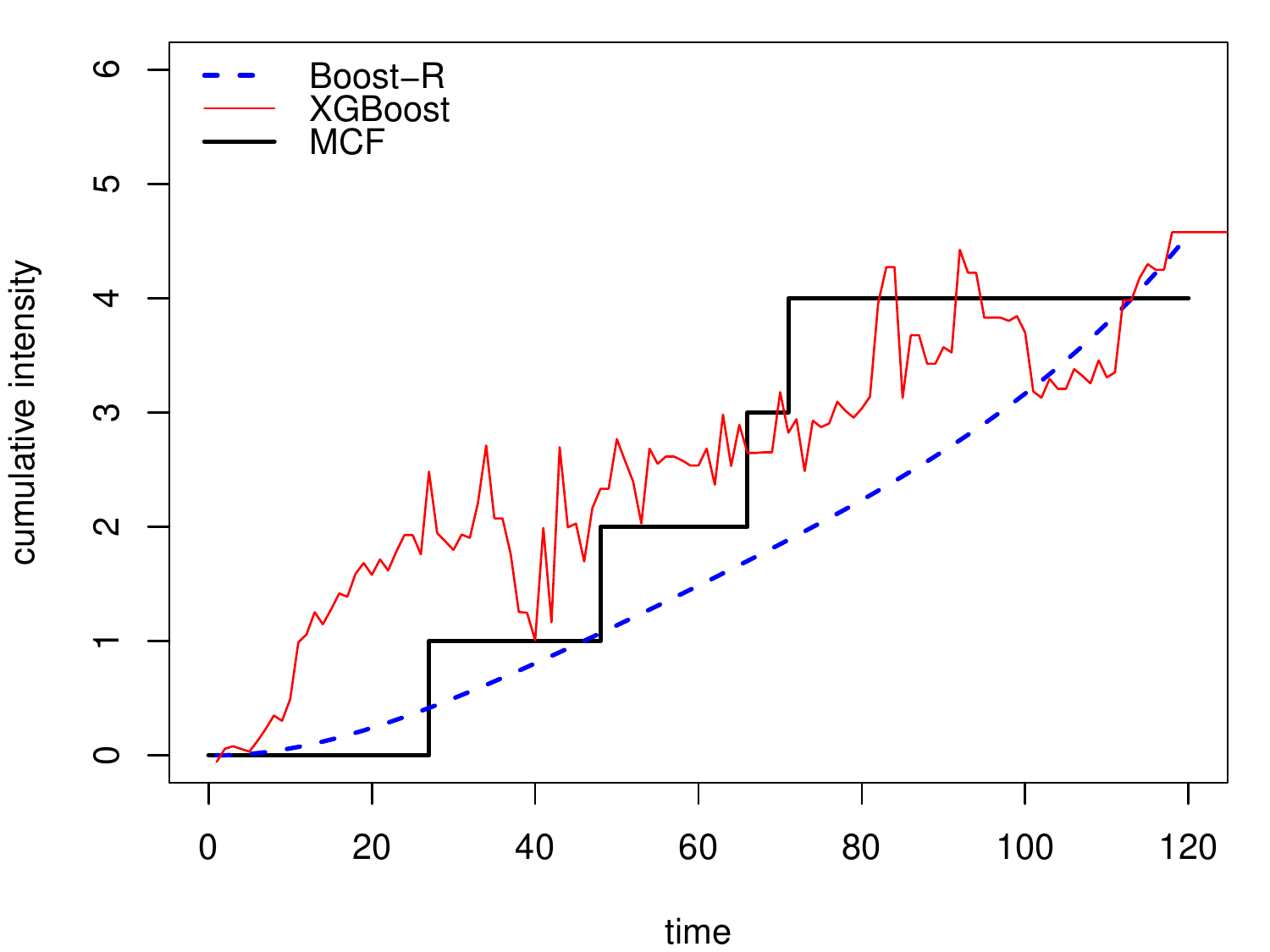}
 		\end{center}
 		\caption{An example that shows how the predicted cumulative intensity from XGBoost and the proposed Boost-R typically look like. The output from XGBoost appears to be bumpy because it models the cumulative events at different times independently.}
 		\label{fig:bumpy}
 	\end{figure} 
 	
 	$\triangleright$ More importantly, if the training data is censored in time (say, at time $t_c$), the \textit{ad hoc} use of XGBoost (which treats time $t$ as another feature) is incapable of event predictions at a time beyond $t_c$ (i.e., extrapolation in time). This is due to the non-parametric nature of tree-based methods which are ineffective in making extrapolations outside the range of the feature in the training dataset. As a result, although XGBoost can learn the relationship between the cumulative intensity and time (treated as a feature) over the interval $[0,t_c]$, the tree-based method cannot predict the future event process over a time interval beyond the censoring time $t_c$ (i.e., outside the range of ``time'' in the training dataset).
 	
 	As clearly shown in the Figure \ref{fig:extrapolation} below (more details are provided in the Appendix of the revised manuscript), the predicted cumulative intensity remains a constant if XGBoost is used with time being treated as a feature. In other words, such a use of XGBoost (after treating time $t$ as another feature) is incapable of making predictions beyond the censoring $t_c$. In the proposed Boost-R, however, because each terminal node contains a time-dependent function (rather than a constant), the predicted cumulative intensity function from Boost-R is smooth and we are able to make extrapolations in time.

 	\begin{figure}[h!]
 		\begin{center}
 			\includegraphics[width=0.7\textwidth]{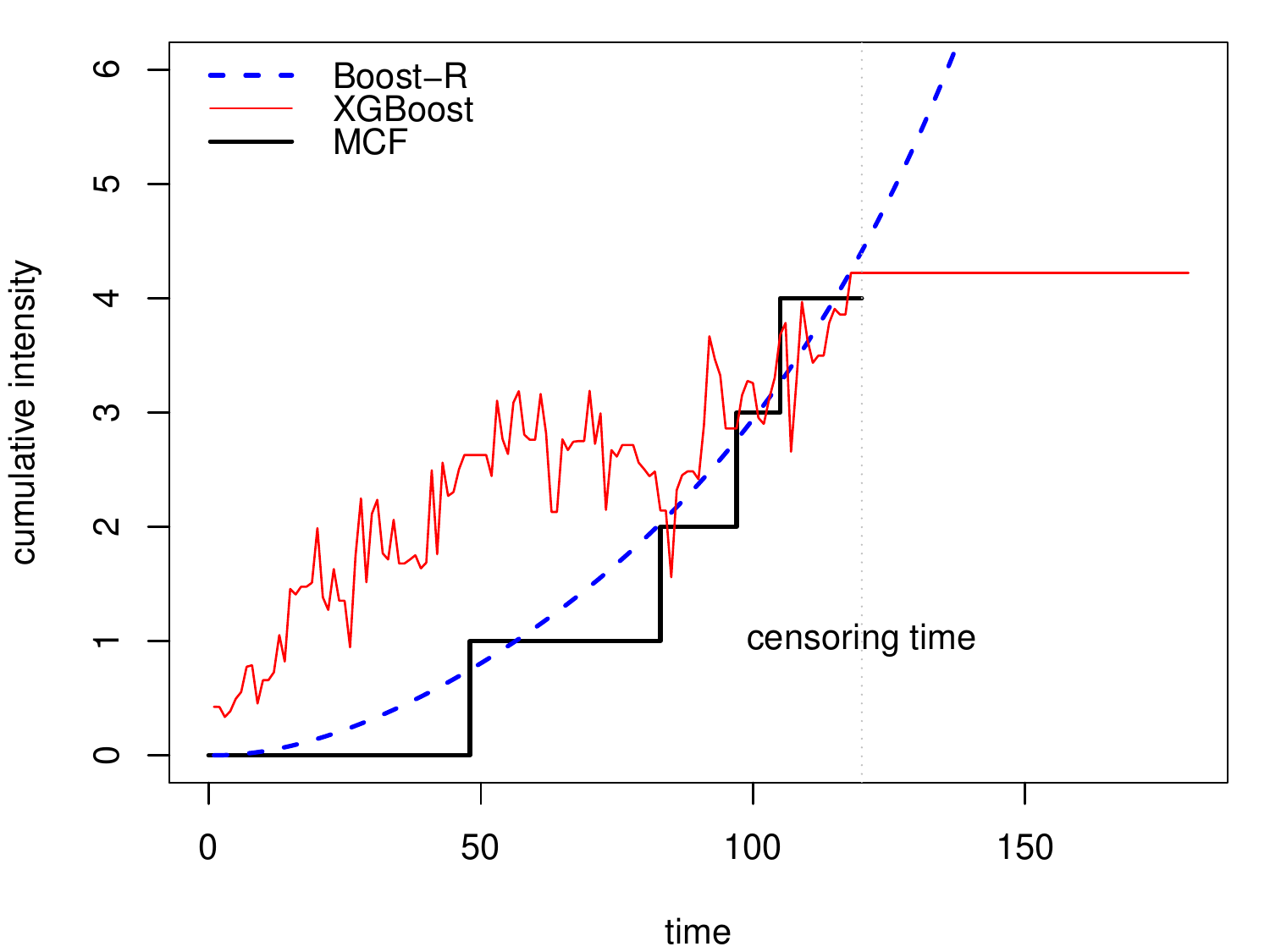}
 		\end{center}
 		\caption{The ad hoc use of XGBoost is not capable of predicting the event process at a future time interval beyond the censoring time (i.e., extrapolation in time).}
 		\label{fig:extrapolation} 
 	\end{figure} 
 	
 	Similarly, if the training data is under Type-II censoring, say, the observation of the process is stopped after observing $n_c$ number of events, the \textit{ad-hoc} use of XGBoost cannot generate predictions that go beyond $n_c$. For example, after training the XGBoost with a training data under Type-II censoring (censored at $n_c$), the model cannot answer common questions such as when the $(n_c+1)$th failure will occur. This is because  no sample in the training dataset has more than $n_c$ failures under Type-II censoring, and the non-parametric nature of the conventional tree-based methods prevents the algorithm from making predictions beyond the largest number of failures in the training dataset.

 	\vspace{32pt}
 \section{Additional Insights on Boost-R and Comparison Studies}
 
 [*Appendix B can be moved to supplementary materials if necessary] \\
 
We provide additional discussions and comparison studies of the proposed Boost-R using another application example. 
 
 \textbf{\textit{Application.}} In \cite{kelly2000}, the authors investigated the recurrent event data modeling for childhood infectious acute respiratory illness (ARI). The goal of the study was to understand the effect of childhood MORbidity from supplementation of VITamin A---the MORVITA trial. This was a randomized double-blinded placebo-controlled trial with 1405 subjects aged 6-47 months. Once a child was randomized (to vitamin A or placebo) they received the same treatment throughout the study. 
 
Each subject has a maximum of four events (i.e., Type-II censoring), and the events are censored if the additive total time since the start of the study is greater than 120 days (i.e., Type-I censoring). 
 
 \vspace{8pt}
  \textbf{\textit{Model.}} \cite{kelly2000} considered a random-effect model. For a subject $i$, let $\lambda_{ik}$ be the intensity between the $(k-1)$th event and the $k$th event ($k=1,2,3,4$), and 
  \begin{equation} \label{eq:A_1}
  \log\lambda_{ik}(t) = \beta_0 + \beta_k Z_{ik} + v_i
  \end{equation}
  where $\beta_0$ and $\beta_k$ are the effects,  $Z_{ik}=1$ if the subject receives treatment after the $(k-1)$th event otherwise $Z_{ik}=0$, and $v_i \sim N(0,\sigma^2)$ is a random effect covariate that introduces the within-subject correlation. 
 
  Different subjects respond to the treatment differently. For example, if the treatment is constantly effective, $\beta_1=\beta_2=\beta_3=\beta_4=-1$. If the treatment is only effective for the first event, $\beta_1=-1$ and $\beta_2=\beta_3=\beta_4=0$. 
  
  \vspace{8pt}
  \textbf{\textit{Data.}} In our experiment, we simulate the data for 1000 subjects based on the model in \cite{kelly2000}. For each subjects, we randomly simulate two features $x_1$ and $x_2$ from a uniform distribution on $[0,1]$, and consider four potential sub-populations as follows
  \begin{itemize}
  	\item if $x_{i,1},x_{i,2} < 0.5$, the treatment is only effective for the first event, i.e., $\beta_1=-1$ and $\beta_2=\beta_3=\beta_4=0$; 
  	\item if $x_{i,1}<0.5$ and $x_{i,2} \geq 0.5$, the treatment is effective for the first two events, i.e., $\beta_1=\beta_2=-1$ and $\beta_3=\beta_4=0$;
  	\item if $x_{i,1}\geq0.5$ and $x_{i,2} < 0.5$, the treatment is effective for the first three events, i.e., $\beta_1=\beta_2=\beta_3=-1$ and $\beta_4=0$.
  	\item if $x_{i,1}\geq0.5$ and $x_{i,2}\geq0.5$, the treatment is effective for the all four events, i.e., $\beta_1=\beta_2=\beta_3=\beta_4=-1$.
  \end{itemize}
  
  Hence, different subjects respond to treatment differently. Even for subjects from the same sub-group, the random effect, $v_i\sim N(0,\sigma^2)$, further introduces the within-sample correlation. In the subsequent comparison studies, we consider three different values for $\sigma$, i.e., $0$, $0.1$ and $0.4$, which were also considered in \cite{kelly2000}.
  
  All datasets are available on GitHub (\url{https://github.com/dnncode/Boost-R}). 
   
  \vspace{8pt}
  \textbf{\textit{Comparison.}} Three methods are included in the comparison study: Boost-R, RF-R and XGBoost (described in Appendix A). 
  For each method, data from 500 subjects are used to train the model, and data from the remaining 500 subjects are used to test the model performance. Figures \ref{fig:comparison_sigma0}, \ref{fig:comparison_sigma01} and \ref{fig:comparison_sigma04} below show the box plot of the squared $L^2$ distance between the predicted cumulative intensity and the observed cumulative intensity over the time interval from 0 to 120 days, respectively for three datasets assuming different values for $\sigma$, i.e., $0$, $0.1$ and $0.4$. 
  
  In each figure, four combinations of the tuning parameters ($\gamma_1$ and $\gamma_2$) are used for Boost-R models. In particular, Boost-R-1, Boost-R-2, Boost-R-3 and Boost-4 are respectively based on the following combinations of $\gamma_1$ and $\gamma_2$: $(10,10)$, $(10,50)$, $(50,10)$ and $(50,50)$. Three different choices of the learning rate, including $0.1$, $0.5$ and $1$, are respectively used for XGBoost-1, XGBoost-2 and XGBoost-3. RF-F does not involve major tuning parameters. 
  
  Figures \ref{fig:comparison_sigma0}, \ref{fig:comparison_sigma01} and \ref{fig:comparison_sigma04} all indicate that the proposed Boost-R provides the best performance for all three datasets. In fact, considering the additional limitations of the ad-hoc use of XGBoost (discussed in Appendix A), Boost-R appears to be a good choice for such an application. In addition, the performance of all methods deteriorates when $\sigma$ becomes larger, as expected. 

  	\begin{figure}[h!]
  	\begin{center}
  		\includegraphics[width=0.9\textwidth]{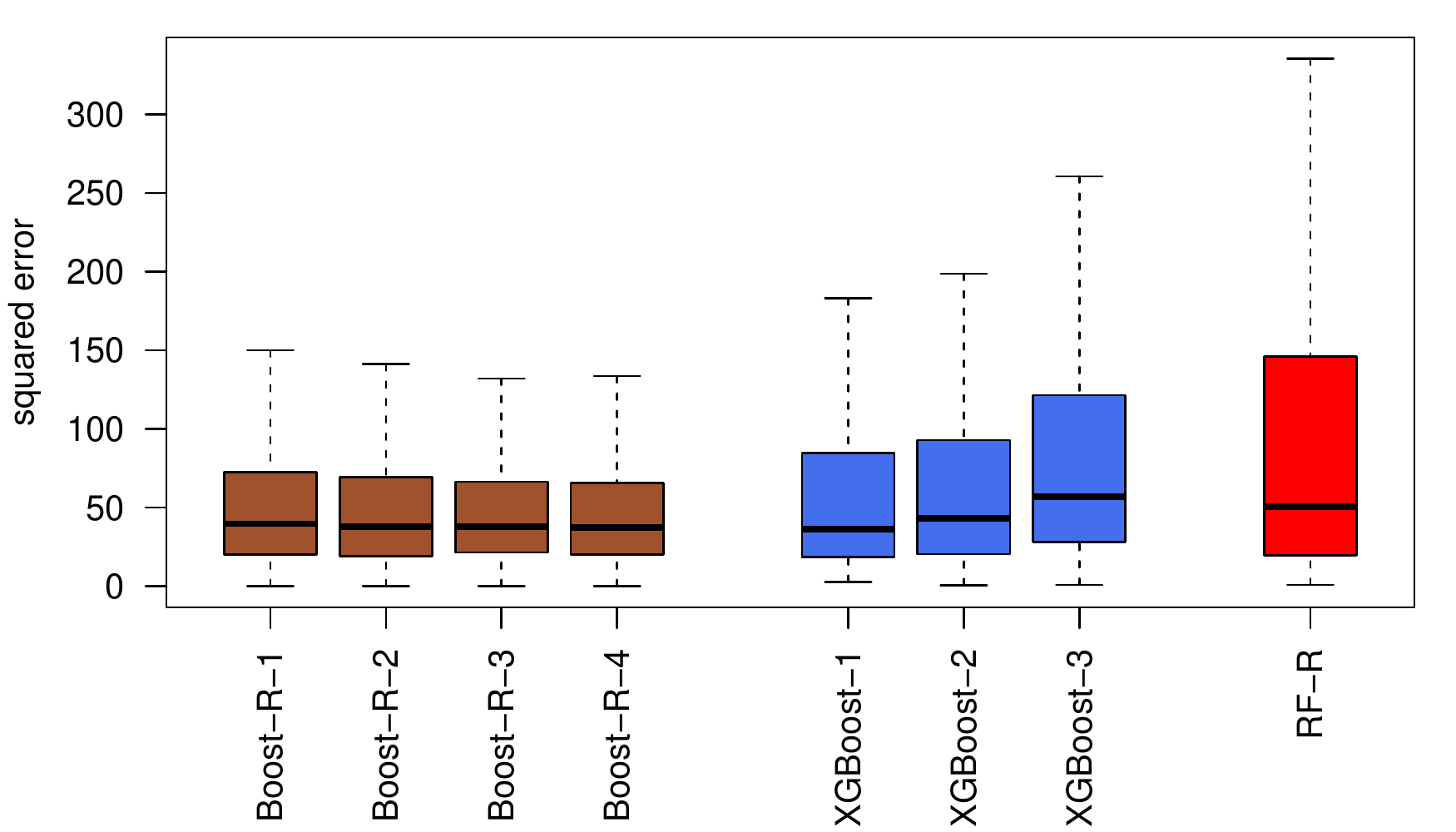}
  	\end{center}
  	\caption{Comparison between Boost-R, RF-R and XGBoost: box plot of the squared $L^2$ distance between the predicted cumulative intensity and the observed cumulative intensity over 0 to 120 days ($\sigma=0$)}
  	\label{fig:comparison_sigma0} 
  \end{figure} 
\begin{figure}[h!]
	\begin{center}
		\includegraphics[width=0.9\textwidth]{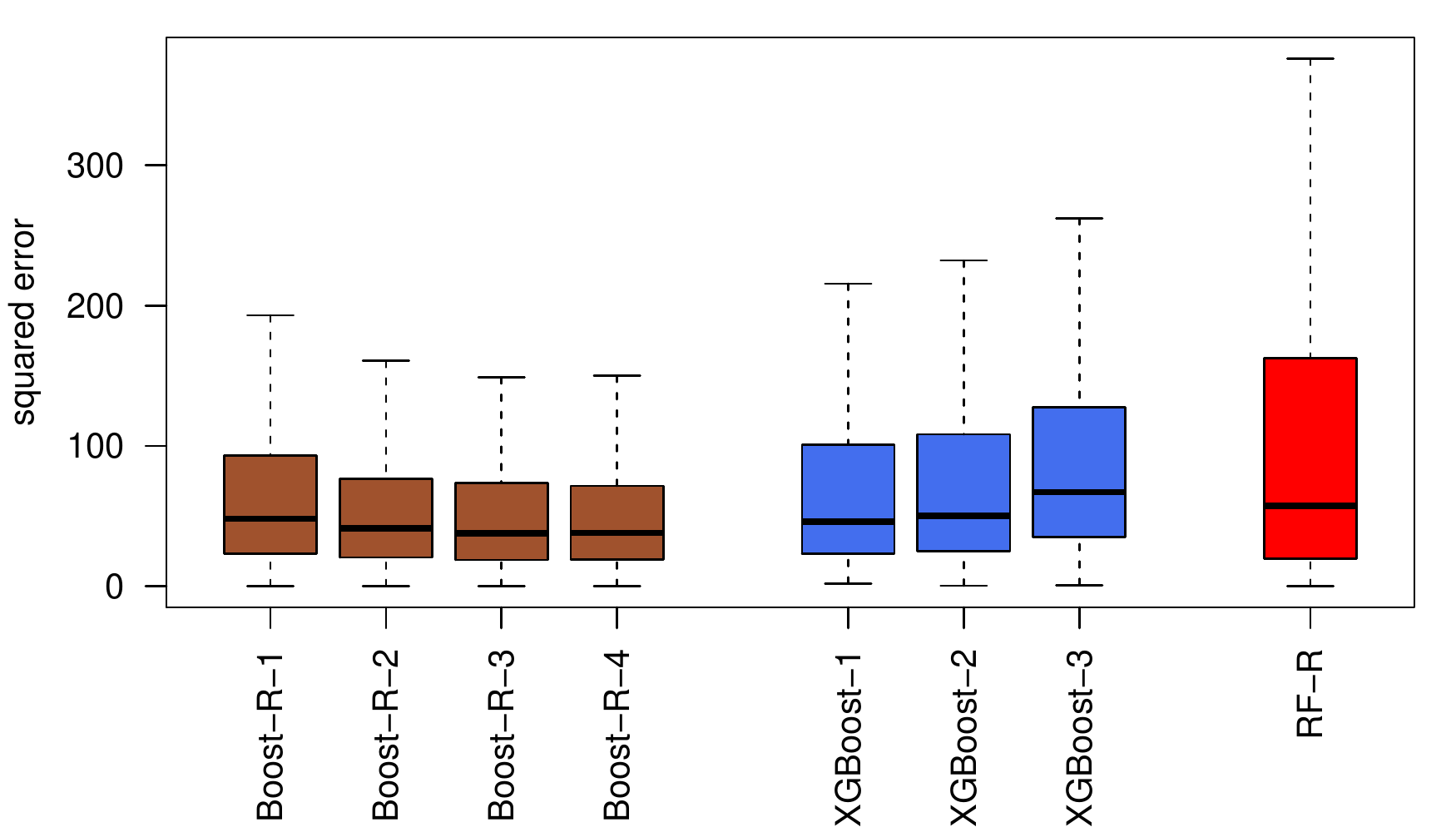}
	\end{center}
	\caption{Comparison between Boost-R, RF-R and XGBoost: box plot of the squared $L^2$ distance between the predicted cumulative intensity and the observed cumulative intensity over 0 to 120 days ($\sigma=0.1$)}
	\label{fig:comparison_sigma01} 
\end{figure} 
\begin{figure}[h!]
	\begin{center}
		\includegraphics[width=0.9\textwidth]{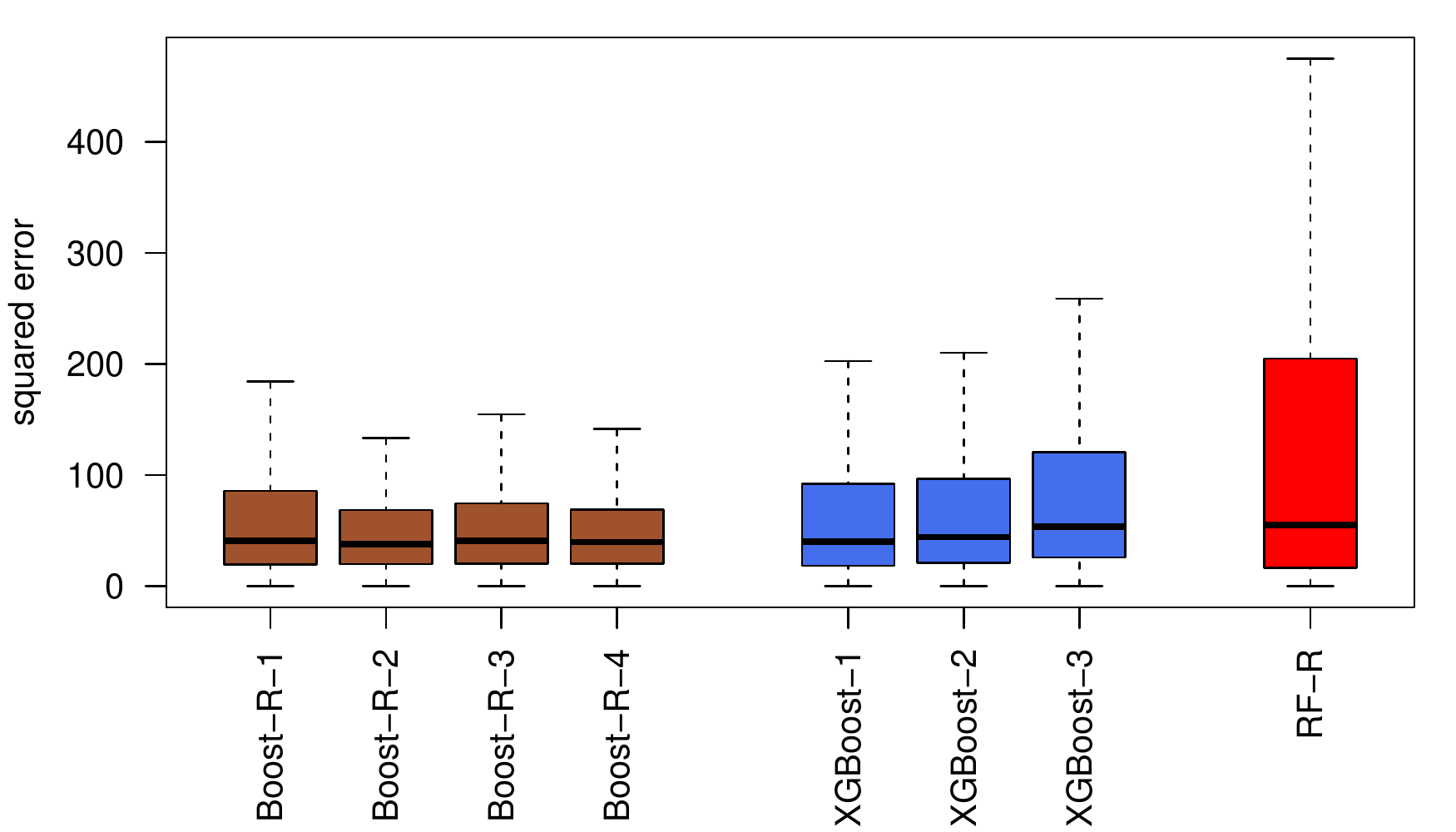}
	\end{center}
	\caption{Comparison between Boost-R, RF-R and XGBoost: box plot of the squared $L^2$ distance between the predicted cumulative intensity and the observed cumulative intensity over 0 to 120 days ($\sigma=0.4$)}
	\label{fig:comparison_sigma04} 
\end{figure}

\clearpage
  We further compare the extrapolation capabilities between Boost-R and the ad-hoc use of XGBoost. In particular, we train the model using the data from 500 subjects over the time interval between 0 and 120 days, and use the model to predict the cumulative number of events at 240 days for the 500 subjects in the testing dataset (i.e., extrapolation in time). Figures \ref{fig:comparison_sigma0_pred}, \ref{fig:comparison_sigma01_pred} and \ref{fig:comparison_sigma04_pred} shows the box plot of the MSE of the predicted cumulative number of events at 240 days. All three figures well illustrate the advantage of Boost-R in terms of extrapolating the number of events beyond the censoring time. Of course, as already discussed in Appendix A, the non-parametric nature of tree-based methods prevents the ad-hoc use of XGBoost to predict the future event process beyond the censoring time (i.e., outside the range of ``time'' in the training dataset).
  
  	\begin{figure}[h!]
  	\begin{center}
  		\includegraphics[width=0.9\textwidth]{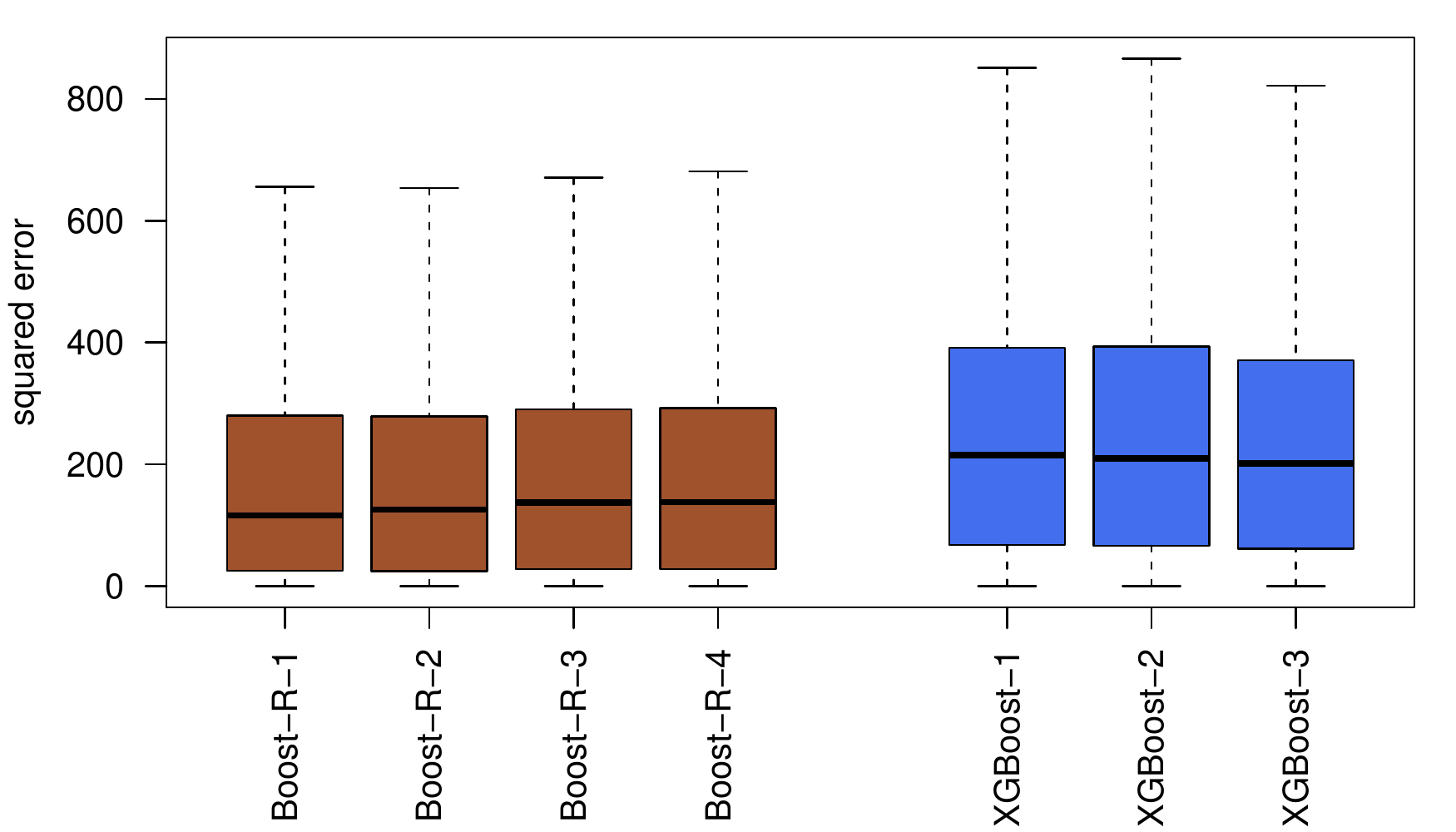}
  	\end{center}
  	\caption{Box plot of the MSE of the predicted cumulative number of events at 240 days ($\sigma=0$)}
  	\label{fig:comparison_sigma0_pred} 
  \end{figure} 
  \begin{figure}[h!]
  	\begin{center}
  		\includegraphics[width=0.9\textwidth]{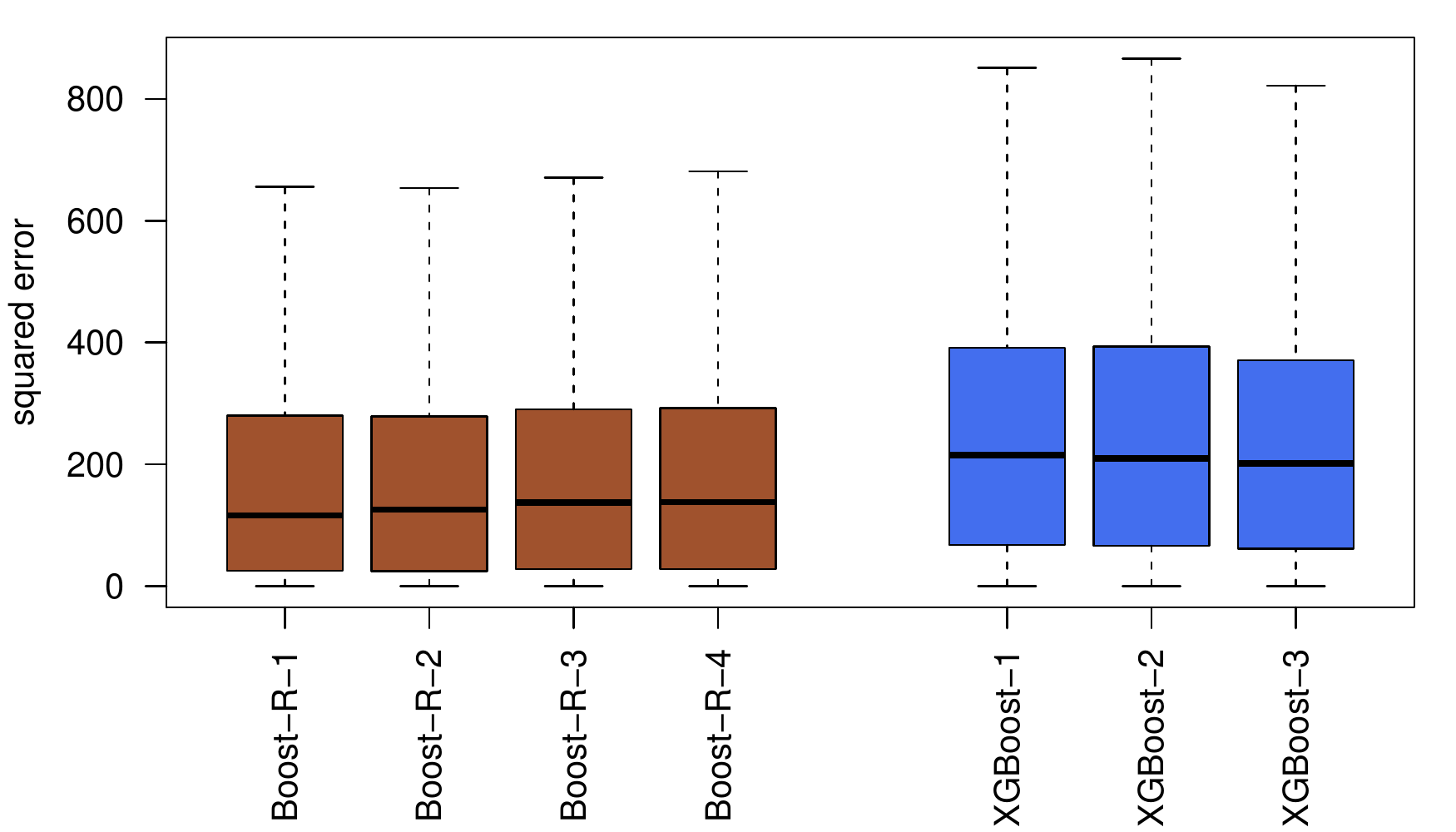}
  	\end{center}
  	\caption{Box plot of the MSE of the predicted cumulative number of events at 240 days ($\sigma=0.1$)}
  	\label{fig:comparison_sigma01_pred} 
  \end{figure} 
  \begin{figure}[h!]
  	\begin{center}
  		\includegraphics[width=0.9\textwidth]{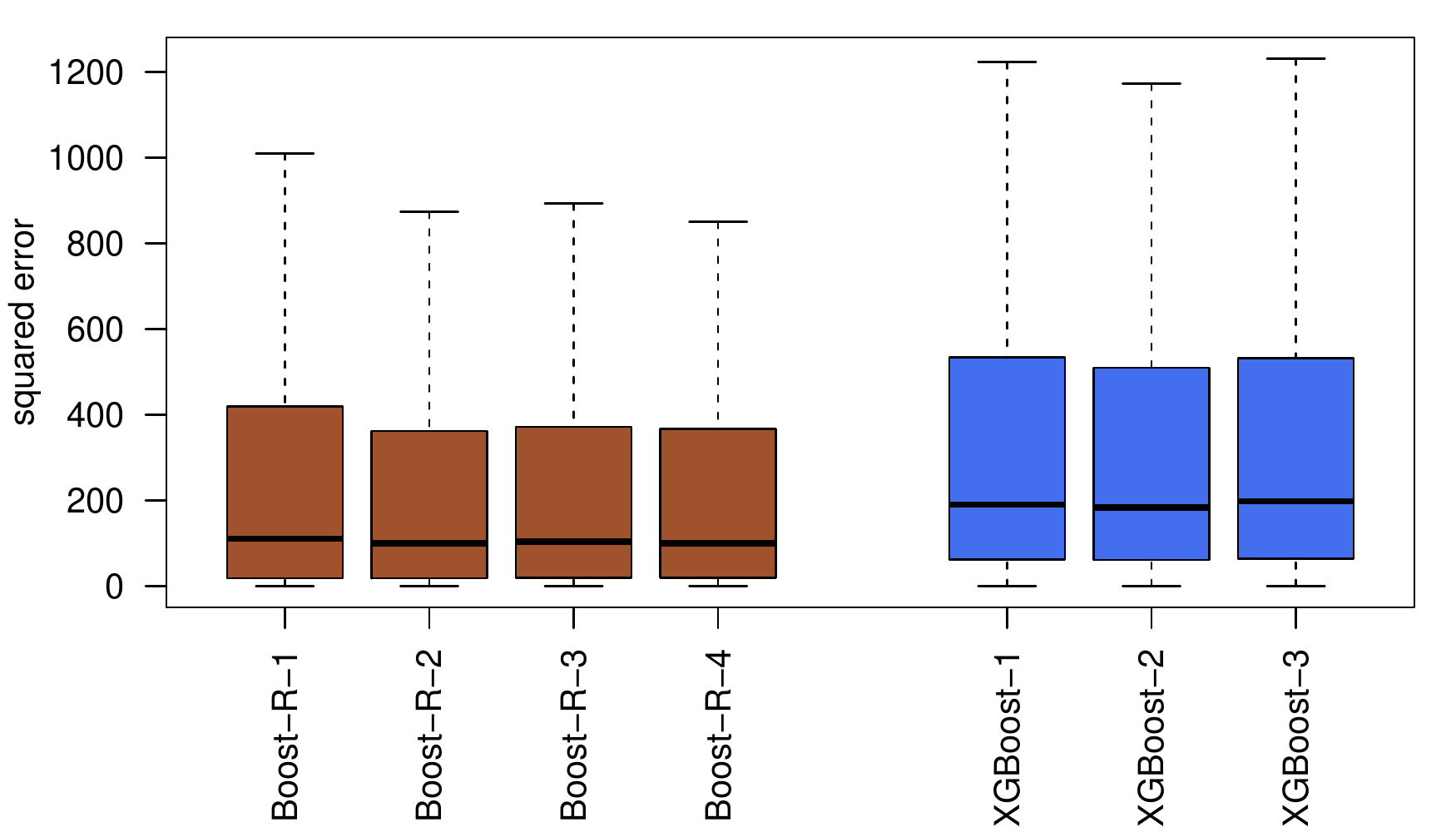}
  	\end{center}
  	\caption{Box plot of the MSE of the predicted cumulative number of events at 240 days ($\sigma=0.4$)}
  	\label{fig:comparison_sigma04_pred} 
  \end{figure}
  
 \end{APPENDICES}




\end{document}